\documentclass[twoside,11pt]{article}

\usepackage{jmlr2e}

\usepackage{amssymb}
\usepackage{lineno}
\usepackage{amsmath}
\usepackage{amssymb}
\usepackage{graphicx}
\usepackage{subfig}
\usepackage{algorithm}
\usepackage{algorithmic}
\usepackage{multirow}
\usepackage{hyperref}
\usepackage{color}
\usepackage{soul}
\usepackage{multirow}

\jmlrheading{1}{XXXX}{X-XX}{X/XX}{XX/XX}{Stephen Tierney, Junbin Gao, Yi Guo and Zhengwu Zhang}

\ShortHeadings{Tractable Clustering of Data on the Curve Manifold}{Tierney, Gao, Guo and Zhang}
\firstpageno{1}

\begin{document}

\title{Tractable Clustering of Data on the Curve Manifold}

\author{\name Stephen Tierney \email stephen.tierney@sydney.edu.au \\
       \AND
       \name Junbin Gao \email junbin.gao@sydney.edu.au \\
       \addr Discipline of Business Analytics, The University of Sydney Business School\\
       University of Sydney\\
       NSW, 2006, Australia
       \AND
       \name Yi Guo \email y.guo@westernsydney.edu.au \\
       \addr Centre for Research in Mathematics, School of Computing, Engineering and Mathematics\\
       University of California\\
       Parramatta, NSW 2150, Australia
       \AND
       \name Zhengwu Zhang \email zzhang@samsi.info \\
       \addr Statistical and Applied Mathematical Sciences Institute\\
       NC 27709-4006, USA}

\editor{XXXXX}

\maketitle

\begin{abstract}
In machine learning it is common to interpret each data point as a vector in Euclidean space. However the data may actually be functional i.e.\ each data point is a function of some variable such as time and the function is discretely sampled. The naive treatment of functional data as traditional multivariate data  can lead to poor performance since the algorithms are ignoring the correlation in the curvature of each function. In this paper we propose a tractable method to cluster functional data or curves by adapting the Euclidean Low-Rank Representation (LRR) to the curve manifold. Experimental evaluation on synthetic and real data reveals that this method massively outperforms prior clustering methods in both speed and accuracy when clustering functional data.
\end{abstract}

\begin{keywords}
Functional Data, Curves, Manifold, Clustering
\end{keywords}

\section{Introduction}

machine learning it is common to interpret each data point as a vector in Euclidean space \citep{Bishop2006}. Such a discretisation is chosen because it allows for easy closed form solutions and fast computation, even with large datasets. However these methods choose to ignore that the data may not naturally fit into this assumption. In fact much of the data collected for practical machine learning are actually functions i.e. curves. In contrast with feature vectors, functional data encodes gradient information, which is vital to analysis. For example financial data such as stock or commodity prices are functions of monetary value over time \citep{maharaj2000cluster}. Recently functional data has become increasingly important in many scientific and engineering research areas such as ECG (electrocardiogram) or EEG (Electroencephalography) in healthcare \citep{kalpakis2001distance}, subject outlines in both macro and micro-biology \citep{lee2004contour, bar2002new, song2008optimal}, weather or climate data \citep{gaffney2004probabilistic}, astronomy \citep{rakthanmanon2012searching} and motion trajectories from computer vision \citep{gaffney2004probabilistic,muller2007dtw}.  

Analyzing functional data has been an emerging topic in statistical research \citep{FerratyRomain2011,Mueller2011,SrivastavaShantanuJermyn2011,SrivastavaWuKurtekKlassenMarron2011} and has attracted great attention from machine learning community in recent years \citep{BahadoriKaleFanLiu2015,PetitjeanForestierWebbNicholsonChenKeogh2014}. One of important challenges in analyzing functional data for machine learning is to efficiently cluster and to learn better representations for functional data. Theoretically the underlying process for functional data is of infinite dimension, thus it is difficult to work with them with only finite samples available.  A desired model for functional data is expected to properly and parsimoniously characterize the nature and variability hidden in the data. The classic functional principal component analysis (fPCA) \citep{RamsaySilverman2005, jacques2014functional} is one of such examples to discover dominant modes of variation in the data. However fPCA may fail to capture patterns if the functional data are not well aligned in its domain.  For time series, a special type of functional data, dynamic time warping (DTW) has long been proposed to compare time series based on shape and distortions (e.g., shifting and stretching) along the temporal axis \citep{Rakthanmanon2013,TuckerWuSrivastava2013}.

Another important type of functional data is shape \citep{SrivastavaShantanuJermyn2011,SuSrivastavaHuffer2013}.   Shape is an important characterizing feature for objects and in computer vision shape has been widely used for the purpose of object detection, tracking, classification, and recognition. In fact, a natural and popular representation for shape analysis is to parametrize boundaries of planar objects as 2D curves. In object recognition, images of the same object should be similar regardless of resolution, lighting, or orientation. Hence an efficient shape representation or shape analysis scheme must be invariant to scale, translation and rotation. A very useful shape representation is the square-root velocity function (SRVF) representation \citep{SrivastavaShantanuJermyn2011,JoshiKlassenSrivastavaJermyn2007}. In general, the resulting SRVF of a continuous shape is square integrable, belonging to the well-defined Hilbert space where appropriate measurement can be applied, refer to Section~\ref{Sec:curve_lrr} for more details. By acknowledging the true nature of the data we can develop more powerful methods that exploit features that would otherwise be ignored or lead to erroneous results with simple linear models. 


Our intention in this study is to consider functional data clustering by accounting for the possible invariance in scaling/stretching, translation and rotation of functional data to help maintain shape characteristics. The focus of this paper is upon functional data where data sets consist of continuous real curves including shapes in Euclidean spaces. More specifically we propose a method of subspace analysis for functional data based on the idea developed in recent subspace clustering. The idea is to apply a feature mapping such as the aforementioned SRVF to the curves so that they are transformed onto the curve manifold, where the subspace analysis can be conducted based on the geometry on the manifold. In particular, we adapt the well known low-rank representation (LRR) framework \citep{LiuLinYanSunYuMa2013} to deal with data that lie on the manifold of open curves by implementing the classical LRR in tangent spaces of the manifold \citep{FuGaoHongTien2015,WangHuGaoSunYin2015,YinGaoGuo2015}. 

The rest of the paper is organized as follows. In section \ref{Sec:related} we discuss related work in this field. In Section \ref{Sec:curve_lrr}, we review the preliminaries about the manifold of open curves and introduce the Curve Low-Rank Representation (cLRR) model. Section \ref{Sec:optimisation} is dedicated to  explaining an efficient algorithm for solving the optimization proposed in cLRR based on the linearized alternative direction method with adaptive penalty (LADMAP) and the algorithm convergence and complexity are also analyzed.  In Section \ref{Sec:experiments}, the proposed model is assessed on both synthetic and real world databases against several state-of-the-art methods. Finally, conclusions are discussed in Section \ref{Sec:conclusions}.

\section{Related Work}
\label{Sec:related}

The simplest approaches to clustering functional data have relied heavily on Dynamic Time Warping (DTW). DTW is an alignment technique which aims to warp the time axis of the data until the difference between the two sequences is minimised \citep{liao2005clustering}. DTW also provides a distance measurement between the two sequences, once aligned. Historically DTW has been mainly used for large scale data mining, where queries are performed to quickly find the nearest neighbours to sequences of data \citep{rakthanmanon2012searching}. However the aligned distance produced by DTW has been used for clustering of functional data. In its simplest form DTW is used to produce a pairwise distance matrix for the entire dataset \citep{keogh1999scaling, jeong2011weighted}. Then the distance matrix is used by a hierarchical or spectral clustering method to produce the final clusters. Although these DTW based approaches are computationally cheap their clustering accuracy leaves much to be desired. This is due to two flaws in DTW. First the distance measurement for DTW is based on a Euclidean distance between each point in the sequence. This totally ignores any gradient based information in the sequence. Second the warping accuracy is dependant on the correct choice of window size. Poor choices of warping window size can have a dramatic impact on warping and alignment accuracy \citep{ratanamahatana2004everything}. Other methods such as DTW-HMM \citep{oates1999clustering} have used DTW based clusters as an initialisation point for more advanced methods such as Hidden Markov Models.

A more sophisticated approach for functional data clustering is to use probabilistic methods \citep{liao2005clustering}. Early methods used simple Gaussian probability models \citep{gaffney2004probabilistic, gaffney2004joint, banfield1993model, mccullagh2009marginal}. However these models only hold for Euclidean vector data where the notions of cluster centres and cluster variance can be easily quantified \citep{zhang2015elastic}. In \citep{zhang2015bayesian} Zhang e.t.\ al proposed Bayesian clustering of curves, which is similar in nature to DTW pairwise distance clustering with advances that address most of the drawbacks. The distance matrix is based on analysis of the data points in their original space i.e.\ the curve manifold. Then the clustering is performed on the distance matrix by using a probabilistic method that simultaneously finds the cluster assignment and the number of clusters automatically.

We now draw the readers attention to subspace analysis methods, Sparse Subspace Clustering (SSC) \citep{ElhamifarVidal2013} and Low-Rank Representation (LRR), as their shared base model is central to our proposed method.
These methods aim to segment the data into clusters with each cluster corresponding to a unique subspace. More formally, given a data matrix of observed column-wise data samples $\mathbf A = [\mathbf{ a_1,a_2,\dots,a_N}] \in \mathbb{R}^{D \times N}$, the objective of subspace clustering is to assign each data sample to its underlying subspace. The basic assumption is that the data within $\mathbf A$ is drawn from a union of $c$ subspaces $\{S_i\}^c_{i=1}$ of dimensions $\{d_i\}^c_{i=1}$. 

The core of both SSC and LRR is to learn an affinity matrix for the given dataset and the learned affinity matrix will be pipelined to a spectral clustering method like nCUT \citep{ShiMalik2000} to obtain the final subspace labels. 
To learn the affinity matrix, SSC relies on the self expressive property \citep{ElhamifarVidal2013}, which is that\begin{quote}
{\it{each data point in a union of subspaces can be efficiently reconstructed by a linear combination of other points in the data}}.
\end{quote}
In other words, each point can be written as a linear combination of the other points i.e.\ $\mathbf{A = A Z}$, where $\mathbf Z \in \mathbb{R}^{N \times N}$ is a matrix of coefficients. Most methods however assume the data generation model $\mathbf{X = A + N}$, where $\mathbf X$ is the observed data and $\mathbf N$ is noise.
Since it is difficult to separate the noise from the data the solution is to relax the self-expressive model to  $\mathbf{X = X Z + E}$, where $\mathbf E$ is a fitting error and is different from $\mathbf N$.

Similarly LRR \citep{LiuLinYanSunYuMa2013} exploits the self expressive property but attempts to learn the global subspace structure by computing the lowest-rank representation of the set of data points. In other words, data points belonging to the same subspace should have similar coefficient patterns. In the presence of noise LRR attempts to minimise the following objective
\begin{align}
\min_{\mathbf{Z, E}} \; \frac{1}{2}\| \mathbf E \|_{\ell} + \textrm{rank}( \mathbf{Z} ), \quad
\text{s.t.} \quad \mathbf{X = XZ + E}.  \label{(1)}
\end{align}
However rank minimisation is an intractable problem. Therefore LRR actually uses the nuclear norm $\| \cdot \|_*$ (sum of the matrix's singular values) as the closest convex relaxation
\begin{align}
\min_{\mathbf{Z, E}} \; \frac{1}{2}\| \mathbf E \|_{\ell} + \| \mathbf{Z} \|_*, \quad
\text{s.t.} \quad \mathbf{X = XZ + E},  \label{(2)}
\end{align}
where $\| \cdot \|_{\ell}$ is a placeholder for the norm most appropriate to the expected noise type. For example in the case of Gaussian noise the best choice is the $\ell_2$ norm i.e.\ $\| \cdot \|_F^2$ and for sparse noise the $\ell_1$ norm should be used.

Both SSC and LRR rely on the linear self expressive property. This property is no longer available in the nonlinear manifold, e.g. the manifold of open curves as mentioned previously. To generalize LRR or SSC for data in the manifold space, we explicitly explore the underlying nonlinear data structure and utilize the techniques of exponential and logarithm mappings to bring data to a local linear space. 

\section{Tractable Clustering over the Manifold of Curves}
\label{Sec:curve_lrr}

As previously discussed, much of the data encountered in real world is functional. In other words it exhibits a curve like structure over a domain. Euclidean linear models are unable to capture the nonlinear invariance embedded in each data point. For example in thermal infra-red data of geological substances a curve may contain a key identifying feature such as a dip near a particular frequency. This dip may shift or vary position over time even for the same substance due to impurities. Under a linear vector model this variation may cause the vector to drastically move in the ambient Euclidean space and cause poor results. Or in other cases the feature may be elongated, shrunk or be subject to some non-uniform warping or scaling. In all these cases the linear approaches such as the aforementioned DTW clustering and LRR will fail to accurately represent the non-linear affinity in the data.

Exploring these unique non-linear invariances in functional data is the focus of this paper. We now discuss how to adapt LRR (also applicable to SSC) such that it easily accepts curve data and the nonlinear relationships within clusters can be easily discovered. This method offers a significant improvement over standard linear LRR since we are computing the similarity of curves using a metric that takes into account the shape (gradient) information. In contrast to probabilistic models such as \citep{zhang2015elastic} that use the raw distance matrix we promote strong block-wise structure through low-rank regularisation over the distance matrix, which grealty improves clustering accuracy. Furthermore by retaining most of the linear model we maintain tractable computational requirements and running times.

\subsection{The Curve Manifold}
Given a smooth parameterized $n$-dimension curve $\beta : D = [0, 1] \to \mathbb R^n$, we represent it using the square-root velocity function (SRVF) representation  \citep{SrivastavaShantanuJermyn2011,JoshiKlassenSrivastavaJermyn2007}, which is given by
\begin{align*}
q(t) = \frac{\dot{\beta}(t)}{\sqrt{\| \dot{\beta}(t) \|}}.
\end{align*}
The SRVF mapping transforms the original curve $\beta(t)$ into a gradient based representation, which facilitates the comparing of the shape information. 

In this paper, we focus on the set of open curves, e.g. the curves do not form a loop ($\beta(0) \neq \beta(1)$). For handling general curves, we refer readers to  \citep{SrivastavaShantanuJermyn2011}. The SRVF facilitates a measure and geometry bearing invariance to scaling, shifting and reparameterization in the curves domain. For example, all the translated curves from a curve $\beta(t)$ will have the same SRVF.  Robinson \citep{Robinson2012} proved that if the curve $\beta(t)$ is absolutely continuous, then its SRVF $q(t)$ is square-integrable, i.e., $q(t)$ is in a functional Hilbert space $L^2(D, \mathbb{R}^n)$ .  Conversely for each $q(t)\in L^2(D, \mathbb{R}^n)$, there exists a curve $\beta(t)$ whose SRVF corresponds to $q(t)$. Thus the set $L^2(D, \mathbb{R}^n)$ is a well-defined representation space of all the curves.  The most important advantage offered by the SRVF framework is that the natural and widely used $L^2$-measure on $L^2(D, \mathbb{R}^n)$ is invariant to the reparameterization. That is, for any two SRVFs $q_1$ and $q_2$ and an arbitrarily chosen reparametrization function (non-decreasing) $t = \gamma(\tau)$, we have 
\[
\|q_1(t) - q_2(t)\|_{L^2} = \|q_1(\gamma(\tau)) - q_2(\gamma(\tau))\|_{L^2}.
\]

This property has been exploited in \citep{BahadoriKaleFanLiu2015} for functional data clustering under the subspace clustering framework. Different from the work proposed in \citep{BahadoriKaleFanLiu2015}, we will adopt the newly developed LRR on manifolds framework to the model of curves LRR, see \citep{FuGaoHongTien2015,WangHuGaoSunYin2015,YinGaoGuo2015}. To see this, we introduce some more notation.  Let $\Gamma$ be the set of all diffeomorphisms from $D=[0,1]$ to $D=[0,1]$. This set collects all the reparametrization mappings. $\Gamma$ is a Lie group with the composition as the group operation and the identity mapping as the identity element. Then all the orbits $[q] = \{ q\circ \gamma = q(\gamma(t)) \;|\; \forall \gamma\in \Gamma\}$ together define the quotient manifold $L^2(D, \mathbb{R}^n)/\Gamma$. 

Without loss of generality, all curves are normalized to have unit length, i.e., $\int_{D}\|\dot{\beta}(t)\|dt = 1$. The SRVFs associated with these curves are elements of a unit hypersphere in the Hilbert space $L^2(D, \mathbb{R}^n)$, i.e., $\int_D \| q(t) \|^2 dt = 1$.  Therefore, under the curve normalization assumption, instead of $L^2(D, \mathbb{R}^n)$, we consider the following unit  hypersphere manifold
\begin{align*}
\mathcal C^o = \bigg\{ q \in L^2(D, \mathbb R^n): \int_D \| q(t) \|^2 dt = 1 \bigg\}.
\end{align*}

The manifold $\mathcal C^o$ has some nice properties, see \citep{AbsilMahonySepulchre2008}.  For any two points $q_0$ and $q_1$ in $\mathcal C^o$, a
geodesic connecting them is given by $\alpha: [0, 1] \rightarrow \mathcal C^o$,
\begin{align}
\alpha (\tau)  = \frac1{\sin(\theta)} (\sin(\theta(1 -\tau))q_0 + \sin(\theta\tau)q_1), 
\end{align}
where $\theta = \cos^{-1}(\langle q_0 , q_1 \rangle)$ is the length of the geodesic. If we take derivative of $\alpha$ w.r.t to $q_1$, the tangent vector
at $q_0$ is
\begin{align}
v = \frac{\theta}{\sin(\theta)}[q_1 - \langle q_0 , q_1 \rangle q_0]. \label{Tangent1}
\end{align} 
The above formula is regarded as the {\it Logarithm} mapping $\log_{q_0} (q_1)$ on the manifold $\mathcal C^o$.

As we are concerned with the shape invariance, i.e., we need to additionally remove the shape-preserving transformations: rotation and curve reparametrization. The manifold concerning us is the quotient space of the manifold $\mathcal S^o = \mathcal C^o/(SO(n) \times \Gamma)$, where $SO(n)$ is the rotation group. Each element $[q]\in \mathcal{S}^o$ is an equivalence class defined by 
\[
[q] = \left\{O q(\gamma(t))\sqrt{\dot{\gamma}(t)}\; |\; O\in SO(n) \text{ and } \gamma \in\Gamma \right\}.
\]

Given any two points $[q_0]$ and $[q_1]$ in $\mathcal S^o$,  a  tangent representative \citep{AbsilMahonySepulchre2008} in the tangent space $T_{[q_0]} (\mathcal S^o)$ can be calculated in the following way, as suggested in \citep{ZhangSuKlassenLeSrivastava2015,SuSrivastava2014} based on \eqref{Tangent1},
\begin{align}
\widetilde{v} = \log_{q_0} (\widetilde{q}_1) = \frac{\widetilde{\theta}}{\sin(\widetilde{\theta})}[\widetilde{q}_1 - \langle q_0 , \widetilde{q}_1 \rangle q_0]. \label{Tangent2}
\end{align}
where $\widetilde{q}_1$ is the representative of $[q_1]$ given by the well-defined algorithm in \citep{ZhangSuKlassenLeSrivastava2015,SuSrivastava2014} and $\widetilde{\theta} = \cos^{-1}(\langle q_0 , \widetilde{q}_1 \rangle)$. In fact, $\widetilde{v}$ is the lifting representation of abstract tangent vector $\log_{[q_0]}([q_1])$ on $T_{[q_0]}(\mathcal{S}^o)$ at $q_1$.

\subsection{The Proposed Curve LRR}
Given a set of $N$ unit-length curves $\{\beta_1(t), ..., \beta_N(t)\}$, denote their SRVFs by $\{q_1(t), ..., q_N(t)\}$ such that $[q_i]\in \mathcal S^o$ and $q_i(t)$ is a representative of the equivalent class $[q_i]$. We cannot apply the standard LRR model \eqref{(2)} directly on the quotient manifold $\mathcal S^o$. This is because \eqref{(2)} indeed relies on the following individual linear combination
\begin{align}
\mathbf x_i = \sum^N_{j=1} z_{ij} \mathbf x_j + \mathbf e_i, \label{linear}
\end{align}
which is invalid for $[q_i]$'s on $\mathcal S^o$.   Note that $z_{ij}$ can be explained as the affinity or similarity between data points $\mathbf{x}_i$ and $\mathbf{x}_j$.

On any manifold, the tangent space at a given point is linearly local approximation to the manifold around the point and the linear combination is valid in the tangent space. This prompts us to replace the affinity relation in \eqref{linear} by the following 
\begin{align}
\log_{[q_i]}([q_i]) = \sum_{j=1}^N w_{ij} \log_{[q_i]} ([q_j]) + \mathbf{e}_i\label{tangentLinear}
\end{align}
with the constraint $\sum_{j=1}^N w_{ij} = 1, i = 1, 2, \dots, N$ to maintain consistency at different locations. The meaning of $w_{ij}$ in \eqref{tangentLinear} is the similarity between curves $\beta_i(t)$ and $\beta_j(t)$ via the ``affinity'' between tangent vectors $\log_{[q_i]}([q_i]) $ and $\log_{[q_i]}([q_j])$ at the first order approximation accuracy.  Each $\log_{[q_i]}([q_j])$ can be calculated by \eqref{Tangent2} and it is obvious that $\log_{[q_i]}([q_i])=0$ for any $i$.

%
%

With all the ingredients at hand, we are fully equipped to propose the curve LRR (cLRR) model as follows 
\begin{align}
\begin{aligned}
\min_{\mathbf W} \lambda \| \mathbf W \|_* + \sum_{i = 1}^N \frac{1}{2} \| \sum_{j=1}^N w_{ij} \log_{[q_i]} ([q_j]) \|_{[q_i]}^2, \\
\textrm{s.t.} \; \sum_{j=1}^N w_{ij} = 1, i = 1, 2, \dots, N.
\end{aligned}\label{Model}
\end{align}
where $\|\cdot\|_{[q]}$ is the metric defined on the manifold, which is defined by the classic $L^2$ Hilbert metric on the tangent space.  

Denote $\mathbf w_i$ the $i$-th row of matrix $\mathbf W$ and define
\begin{align} 
\label{b_create}
B^i_{jk} = \langle \log_{[q_i]} ([q_j]), \log_{[q_i]} ([q_k]) \rangle.
\end{align}
Then with some algebraic manipulation we can re-write the model \eqref{Model} into the following simplified form, 
\begin{align}
\begin{aligned}
\min_{\mathbf W} \lambda \| \mathbf W \|_* + \sum_{i = 1}^N \mathbf w_i \mathbf B^i \mathbf w_i^T, \\
\textrm{s.t.} \; \sum_{j=1}^N w_{ij} = 1, i = 1, 2, \dots, N.
\end{aligned}\label{obj_curve}
\end{align}
where $\mathbf B^i  = (B^i_{jk})$.
 
Effectively this objective allows for similarity between curves to be measured in their tangent spaces. Our highly accurate segmentation results in Section \ref{Sec:experiments} have demonstrated that this is an effective way to learn  non-linear similarity.

\section{Optimisation}
\label{Sec:optimisation}
\subsection{Algorithm}
To solve the cLRR objective we use the Linearized Alternative Direction Method with Adaptive Penalty (LADMAP) \citep{LinLiuLi2015,LinLiuSu2011}. First take the Augmented Lagrangian of the objective \eqref{obj_curve}
\begin{equation}
\begin{aligned}
L = &\lambda \|\mathbf W\|_*  + \frac12\sum^N_{i=1}\mathbf w_i \mathbf{B}^i\mathbf w^T_i + \langle {\mathbf y}, \mathbf W\mathbf 1 - \mathbf 1\rangle \\
&+ \frac{\beta}2\|\mathbf W\mathbf 1 - \mathbf 1\|^2_F
\end{aligned}\label{25August2014-5}
\end{equation}
where $\mathbf y$ is the Lagrangian multiplier (vector) corresponding to the equality constraint $\mathbf W\mathbf 1 = \mathbf 1$, $\|\cdot\|_{F}$ is the matrix Frobebius-norm, and we will update $\beta$ as well in the iterative algorithm to be introduced.

Denote by $F(\mathbf W)$ the function defined by \eqref{25August2014-5} except for the first term $\lambda \|\mathbf W\|_*$. To solve \eqref{25August2014-5}, we adopt a linearization of $F(\mathbf{W})$ at the current location $\mathbf W^{(k)}$ in the iteration process, that is, we approximate $F(\mathbf W)$ by the following linearization with a proximal term
\begin{align*}
F(\mathbf W)\approx & F(\mathbf W^{(k)}) + \langle \partial F(\mathbf W^{(k)}), \mathbf W-\mathbf W^{(k)}\rangle \\
&+ \frac{\eta_{W}\beta_k}2\|\mathbf W-\mathbf W^{(k)}\|^2_F,
\end{align*}
where $\eta_W$ is an approximate constant with a suggested value given by $\eta_W = \max\{\|B_i\|^2\}+N+1$, and $\partial F(\mathbf W^{(k)})$ is a gradient matrix of $F(\mathbf W)$ at $\mathbf W^{(k)}$. Denote by $\mathbf B$ the 3-order tensor whose $i$-th front slice is given by $\mathbf B^i$. Let us define $\mathbf W\odot \mathbf B$ the matrix whose $i$-row is given by $\mathbf w_i \mathbf B^i$, then it is easy to show
\begin{equation} 
\partial F(\mathbf W^{(k)}) = \mathbf W\odot \mathbf B + \mathbf y\mathbf 1^T + \beta_k (\mathbf W\mathbf 1 - \mathbf 1)\mathbf 1^T.
\label{Eq:14October2014-4}
\end{equation}

Then \eqref{25August2014-5} can be approximated by linearization and $\mathbf w$ will be updated by the following
\begin{align}
 \mathbf W^{(k+1)}   
= &\arg\min_{\mathbf W} \lambda \|\mathbf W\|_* \label{SolutionW}\\
+&\frac{\eta_W\beta_k}{2} \bigg\|\mathbf W - \left(\mathbf W^{(k)} - \frac{1}{\eta_W\beta_k} \partial F(\mathbf W^{(k)})\right)\bigg\|^2_F. \notag
\end{align}

\begin{algorithm}[]
\caption{{\bf Solving \eqref{obj_curve} by LADMAP}}
\label{curve_lrr_alg}
\begin{algorithmic}[1]

\REQUIRE $\{\mathbf X_i\}_{i=1}^N$, $\lambda$

\STATE Initialise: $\mathbf W = \mathbf 0$, $\mathbf y = \mathbf 0$, $\beta = 0.1$, $\beta_{\text{max}} = 10$, $\rho^0 = 1.1$, $\eta = \max \{ \| \mathbf B^i \|_F \} + N + 1$, $\epsilon_1 = 1e^{-4}$, $\epsilon_2 = 1e^{-4}$

\STATE Construct each $\mathbf B^i$ as per \eqref{b_create}

\WHILE{not converged}

\STATE Update $\mathbf W$ using \eqref{SolutionWk}

\STATE Check convergence criteria
\begin{align*}
\beta^{(k)} \| \mathbf W^{(k+1)} - \mathbf W^{(k)} \|_F \leq \epsilon_1\\
\| \mathbf W \mathbf 1 - \mathbf 1 \|_F \leq \epsilon_2 \
\end{align*}

\STATE Update Lagrangian Multiplier
\begin{align*}
\mathbf y^{(k+1)} = \mathbf y^k + \beta^{(k)} ( \mathbf W \mathbf 1 - \mathbf 1 )^T
\end{align*}

\STATE Update $\rho$
\begin{align*}
\rho = 
\begin{cases}
\rho_0 & \text{if} \;\; \beta^{(k)} \| \mathbf W^{(k+1)} - \mathbf W^{(k)} \|_F \leq \epsilon_1 \\
1 & \text{otherwise,}
\end{cases}
\end{align*}

\STATE Update $\beta$
\begin{align*}
\beta^{(k+1)} = \textrm{min}(\beta_{\textrm{max}}, \rho \beta^{(k)})
\end{align*}
 
\ENDWHILE

\RETURN $\mathbf W$

\end{algorithmic}
\end{algorithm}

Problem \eqref{SolutionW} admits a closed form solution by using SVD thresholding operator \citep{CaiCandesShen2008}, given by
\begin{equation}\label{SolutionWk}
\begin{aligned}
\mathbf W^{(k+1)} = U_{W} S_{\frac{\lambda}{\eta_W\beta_k}}(\Sigma_{W})V_{W}^{T},
\end{aligned}
\end{equation}
where $U_{W}\Sigma_{W}V_{W}^{T}$ is the SVD of $\mathbf W^{(k)} - \frac{1}{\eta_W\beta_k} \partial F(\mathbf W^{(k)})$ and $S_\tau(\cdot)$ is the Singular Value Thresholding (SVT) \citep{CaiCandesShen2008,parikh2013proximal} operator defined by
\begin{equation}
\begin{aligned}
S_{\tau}(\Sigma) = \text{diag}(\max\{|\Sigma_{ii}|-\tau, 0\}).
\end{aligned}
\end{equation}

The updating rule for $\mathbf y$
\begin{equation} 
\mathbf y^{(k+1)} = \mathbf y^{(k)} + \beta_k (\mathbf W^{(k)}\mathbf 1 -\mathbf 1) 
\label{Eq:14October2014-5}
\end{equation}
and the updating rule for $\beta_k$
\begin{equation}
\begin{aligned}
\beta_{k+1} = \min\{\beta_{\text{max}}, \rho \beta_k\},
\end{aligned}\label{UpdateBeta}
\end{equation}
where
 \[
 \rho = \begin{cases} \rho_0 & \beta_k \|\mathbf W^{k+1} - \mathbf W^k\| \leq \varepsilon_1,\\
 1 & \text{otherwise}.
 \end{cases}
 \]

We summarize the above as Algorithm~\ref{curve_lrr_alg}.  Once the coefficient matrix $\mathbf W$ is found, a spectral clustering like nCUT \citep{ShiMalik2000} is applied on the affinity matrix $\frac{|\mathbf W|+|\mathbf W|^T}{2}$ to obtain the segmentation of the data.

\subsection{Complexity Analysis}
For ease of analysis, we firstly define some symbols used in the following. Let $K$ and $r$ denote the total number of iterations and the lowest rank of the matrix $\mathbf W$, respectively. The size of $\mathbf W$ is $N\times N$. The major computation cost of our proposed method contains two parts, calculating all $\mathbf B^i$'s and updating $\mathbf W$. In terms of the formula \eqref{b_create} through \eqref{Tangent1} and \eqref{Tangent2}, the computational complexity of Log algorithm is $O(T^2)$ where $T$ is the number of terms in a discretized curves; therefore, the complexity of $B_{jk}^i$ is at most $O(T^2)$ and $\mathbf B^i$'s computational complexity is $O(N^2T^2)$. Thus the total for all the $\mathbf B^i$ is $O(N^3)$. In each iteration of the Algorithm, the singular value thresholding is adopted to update the low rank matrix $\mathbf W$ whose complexity is $O(rN^2)$~\citep{LiuLinYanSunYuMa2013}. Suppose the algorithm is terminated after $K$ iterations, the overall computational complexity is given by
\[
O(N^3)+O(KrN^2)
\]

\subsection{Convergence Analysis}
Algorithm~\ref{curve_lrr_alg} is adopted from the algorithm proposed in~\citep{LinLiuSu2011}. However due to the terms of $\mathbf B^i$'s in the objective function~\eqref{25August2014-5}, the convergence theorem proved in~\citep{LinLiuSu2011} cannot be directly applied to this case as the linearization is implemented on both the augmented Lagrangian terms and the term involving $\mathbf B^i$'s. Fortunately we can employ the revised approach, presented in \citep{YinGaoLinShiGuo2015}, to prove the convergence for the algorithm. Without repeating all the details, we present the convergence theorem for Algorithm~\ref{curve_lrr_alg} as follows.
\begin{theorem}[Convergence of Algorithm~\ref{curve_lrr_alg}] If $\eta_W\geq \max\{\|B_i\|^2\}+N+1$, $\displaystyle\sum^{+\infty}_{k=1}\beta^{-1}_k = +\infty$, $\displaystyle\beta_{k+1}-\beta_k > C_0 \frac{\sum_i \|B_i\|^2}{\eta_W - \max\{\|B_i\|^2\}-N}$, where $C_0$ is a given constant and $\|\cdot\|$ is the matrix spectral norm, then the sequence $\{W^{k}\}$ generated by Algorithm~\ref{curve_lrr_alg} converges to an optimal solution to problem~\eqref{obj_curve}.
\end{theorem}

In all the experiments we have conducted, the algorithm converges very fast with $K<100$.

\section{Experimental Analysis}
\label{Sec:experiments}

In this section we evaluate the clustering performance of our newly proposed method, cLRR, on synthetic, semi-synthetic and real world datasets. We compare our algorithm to two baseline algorithms: k-means and spectral clustering of a DTW distance matrix. We also compare against the classical LRR and the state of the art Bayesian Clustering of Curves. 

In an effort to maximise transparency and repeatability, all MATLAB code and data used for these experiments can be found online at \url{https://github.com/sjtrny/curveLRR}.

To help evaluate consistency we fixed the parameters to the same values for every experiment. Parameters were selected by testing a wide range of values over all datasets so that the best average result for each method was obtained. For both cLRR and LRR $\lambda$ was set to $0.1$. Overall we found that the segmentation accuracy of LRR did not vary that much with changes in $\lambda$. For LRR and k-means where multidimensional data cannot be easily dealt with, we simply concatenate each dimension into a single vector for each data point. For our DTW baseline algorithm we set the warping window to $10\%$ of the data length, which has been shown to be suitable in most cases \citep{ratanamahatana2004everything}. 

Segmentation accuracy was measured using the subspace clustering accuracy (SCA) metric \citep{ElhamifarVidal2013}, which is defined as
\begin{align}
\text{SCA} = 1 - \frac{\text{num.\ of misclassified points}}{\text{total num.\ of points}},
\end{align}
where higher SCA $\%$ means greater clustering accuracy. The SCA metric is taken over all possible pairwise assignments of clusters.

\subsection{Toy Synthetic Data Clustering}

\begin{table*}
\centering
\begin{tabular}{c c c c c c c}
\hline
 & Mean & Median & Max & Min & Std & Mean Run Time (s) \\
\hline

kmeans & 55.2\% & 55.0\% & 71.7\% & 35.0\% & 9.4\% & 0.01 \\ 
DTW & 58.0\% & 58.3\% & 80.0\% & 40.0\% & 12.2\% & 0.13 \\ 
LRR & 81.6\% & 84.2\% & 91.7\% & 60.0\% & 8.3\% & 0.06 \\ 
Bayesian & 86.6\% & \hl{96.7\%} & \hl{100.0\%} & 55.0\% & 16.4\% & 53.39 \\ 
CurveLRR & \hl{95.3\%} & \hl{96.7\%} & 98.3\% & \hl{88.3\%} & 3.1\% & 20.30 \\ 

\end{tabular}

\caption{Synthetic Results}
\label{table_syn_results}
\end{table*}

\begin{figure*}[]
\centering
	\subfloat[Cluster 1]{
	\includegraphics[width=0.3\linewidth]{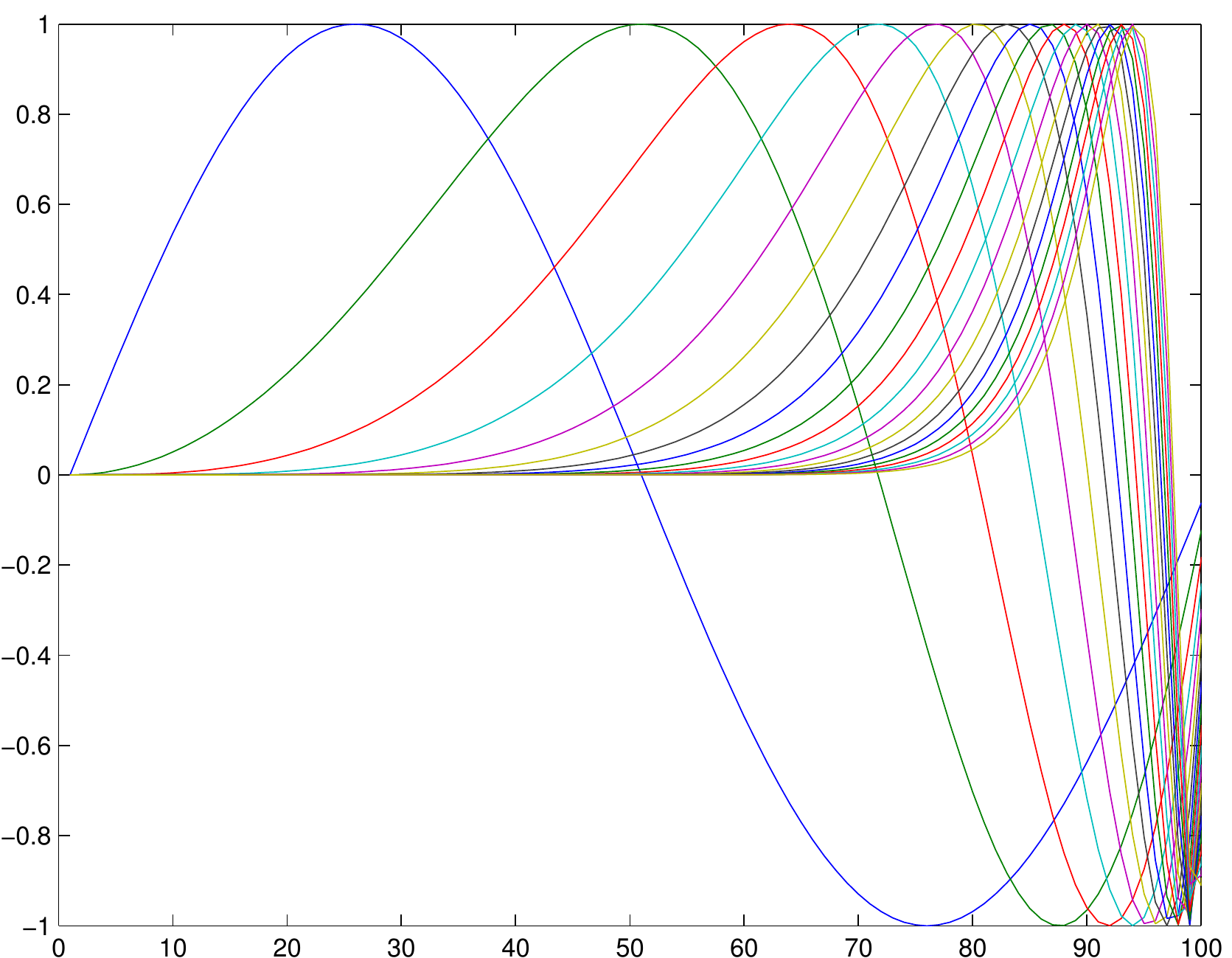}}
	\subfloat[Cluster 2]{
	\includegraphics[width=0.3\linewidth]{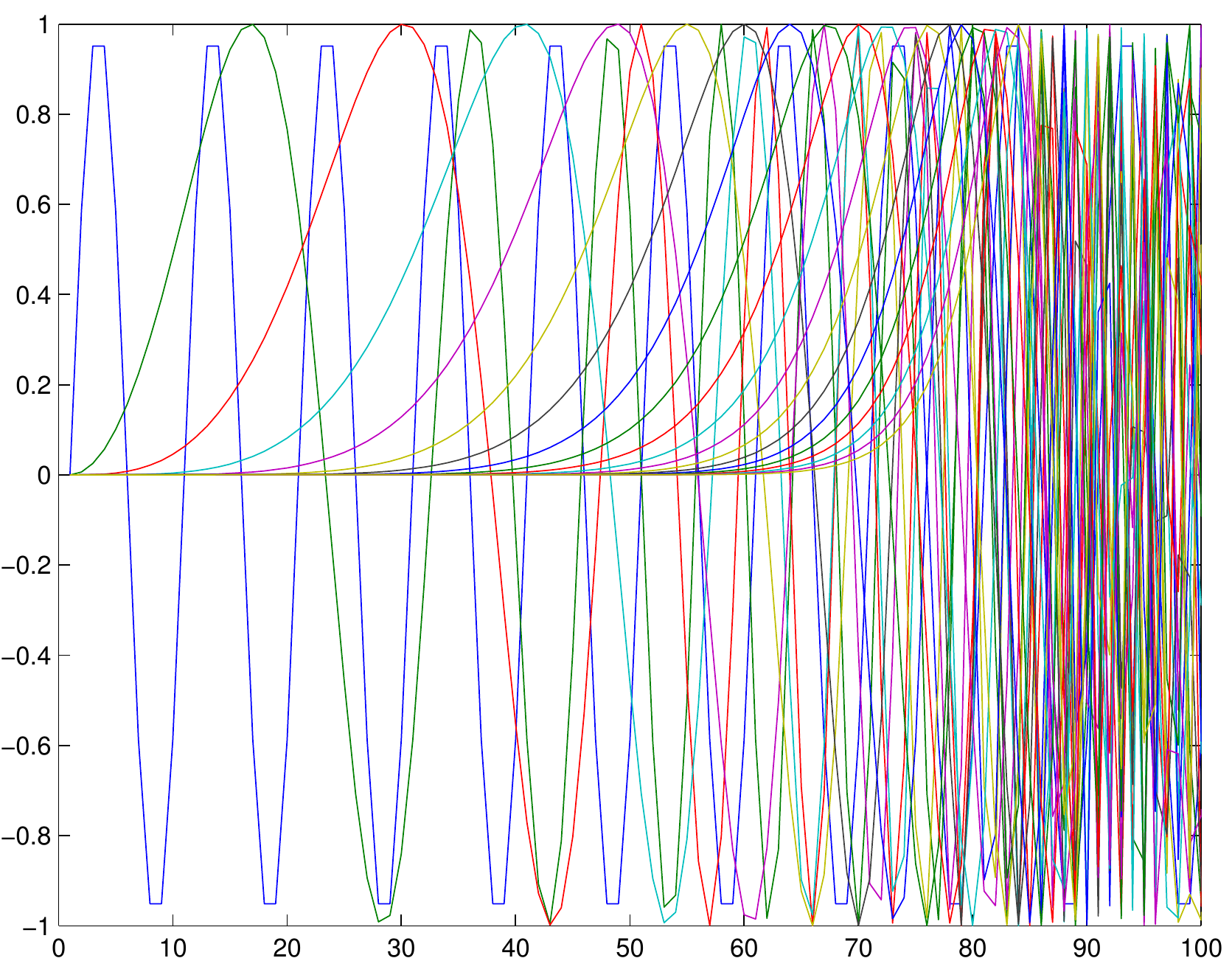}}
	\subfloat[Cluster 3]{
	\includegraphics[width=0.3\linewidth]{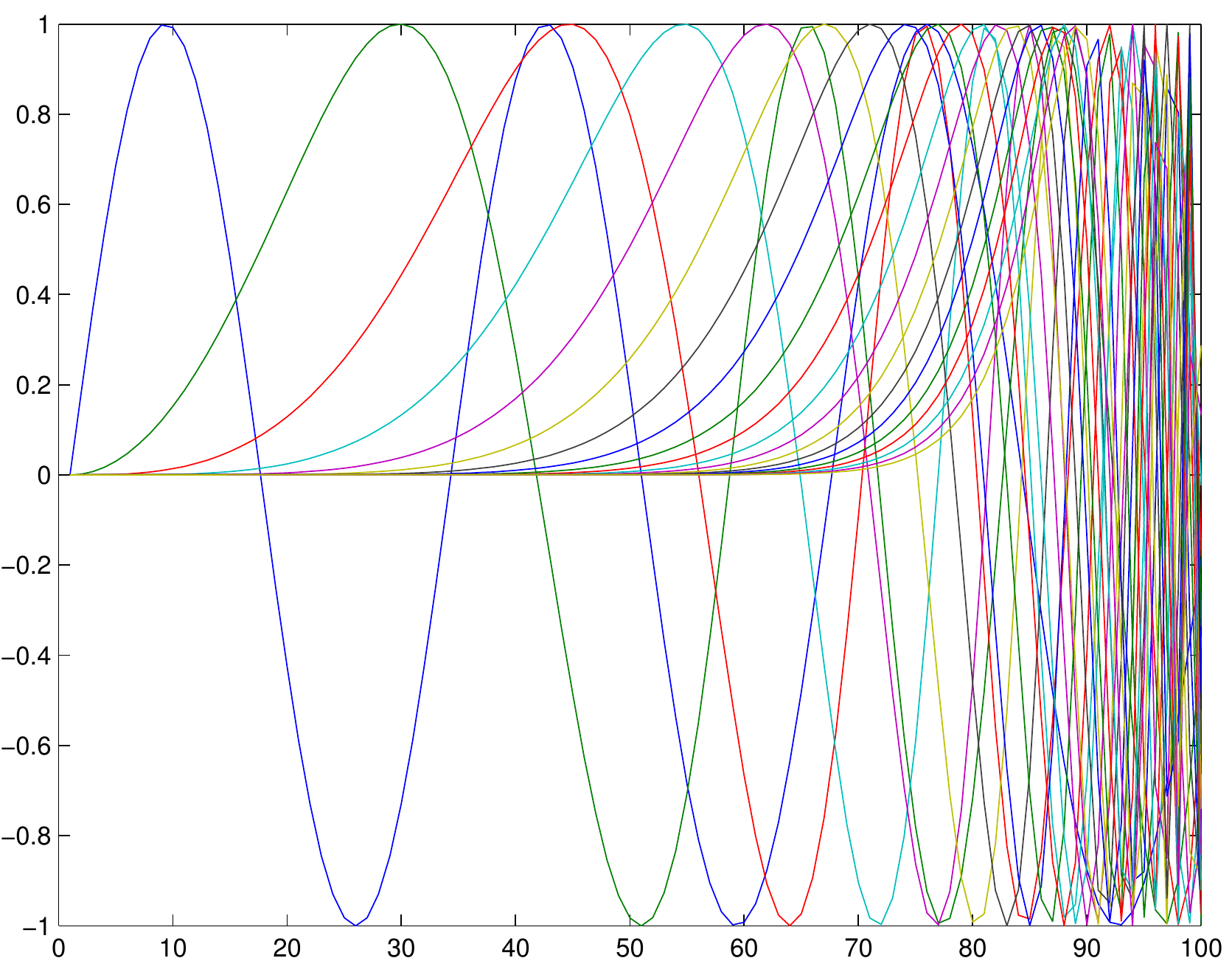}}
	
\caption{Examples of clusters and their curves generated in the Synthetic Data Experiment. Each cluster has a base sine curve (the left most blue curve) which is progressively warped with each successive instantiation.}
\label{fig_syn_data}
\end{figure*}

First we attempt to confirm our hypothesis that cLRR will more accurately capture nonlinear invariance that is otherwise impossible to capture with purely linear or Euclidean distance based methods such as LRR and k-means. In this test three clusters are created consisting of twenty 1-D curves of length 100. The curves in each cluster were sine waves, with each cluster corresponding to a unique frequency. Within each cluster we applied progressive amounts of warping. See Figure \ref{fig_syn_data} for an example of data from three synthetically generated clusters. This experiment was repeated $50$ times with new data generated each time to obtain basic statistics.

Results are reported using subspace clustering accuracy and can be found in Table \ref{table_syn_results}. From the figure it can be clearly observed that this is a challenging dataset since the inter-class variance of the data points is low. However in this experiment cLRR achieves very high clustering accuracy with very low variance in clustering accuracy. Furthermore the run time of cLRR is over half the time of the Bayesian clustering algorithm

\subsection{Semi-Synthetic Thermal Infrared Spectra Clustering}

\begin{table*}
\centering
\begin{tabular}{c c c c c c c}
\hline
 & Mean & Median & Max & Min & Std & Mean Run Time (s) \\
\hline

kmeans & 58.6\% & 58.3\% & 73.3\% & 43.3\% & 6.1\% & 0.01 \\ 
DTW & 65.8\% & 65.8\% & 81.7\% & 46.7\% & 6.0\% & 0.74 \\ 
LRR & 60.2\% & 60.8\% & 66.7\% & 43.3\% & 5.0\% & 0.14 \\ 
Bayesian & 94.7\% & \hl{100.0\%} & \hl{100.0\%} & 66.7\% & 12.3\% & 311.66 \\ 
CurveLRR & \hl{100.0\%} & \hl{100.0\%} & \hl{100.0\%} & \hl{100.0\%} & 0.0\% & 190.50 \\ 

\end{tabular}

\caption{Thermal Infrared Results}
\label{table_tir_results}
\end{table*}

\begin{figure*}[]
\centering
	\subfloat[Cluster 1]{
	\includegraphics[width=0.24\linewidth]{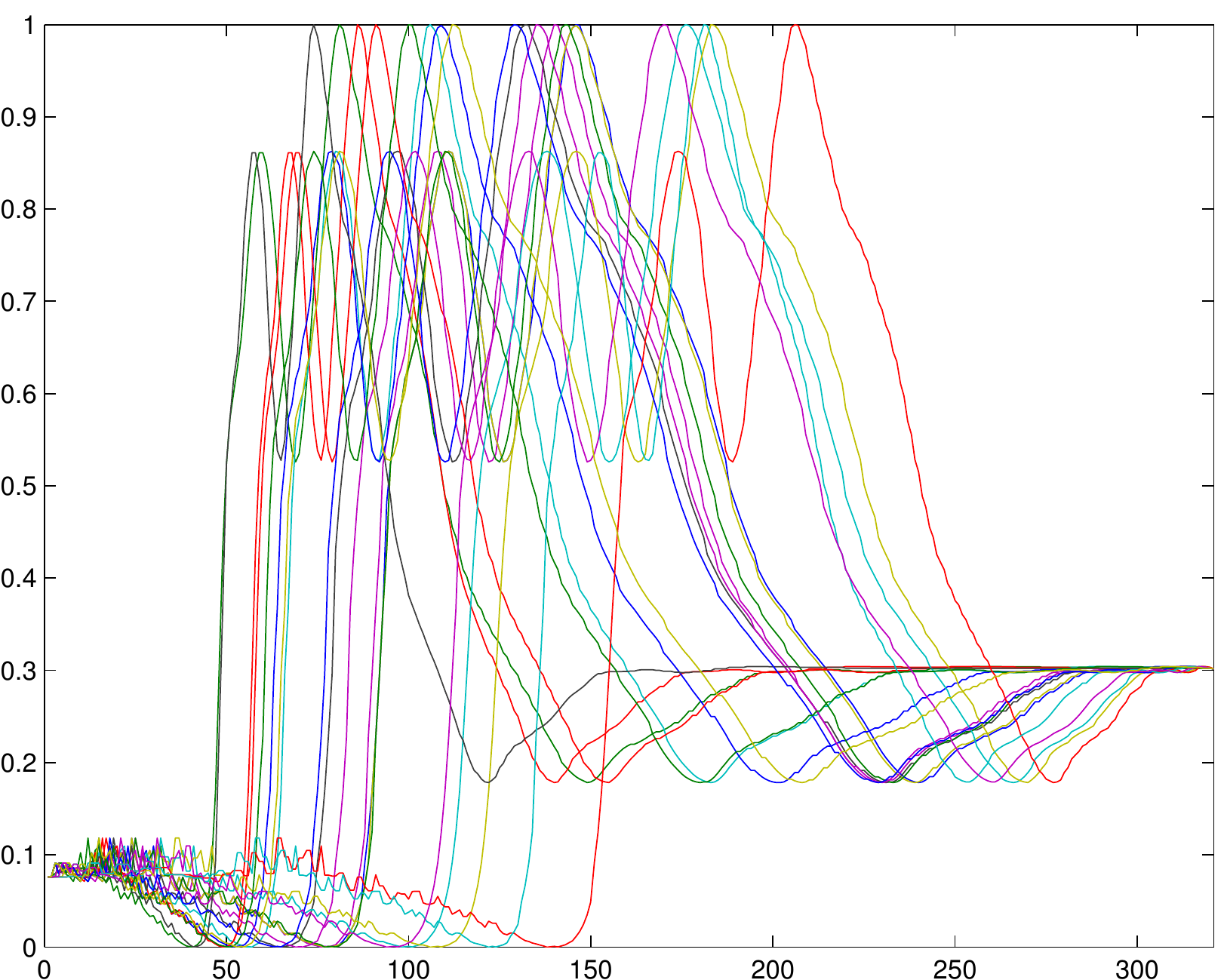}}
	\subfloat[Cluster 2]{
	\includegraphics[width=0.24\linewidth]{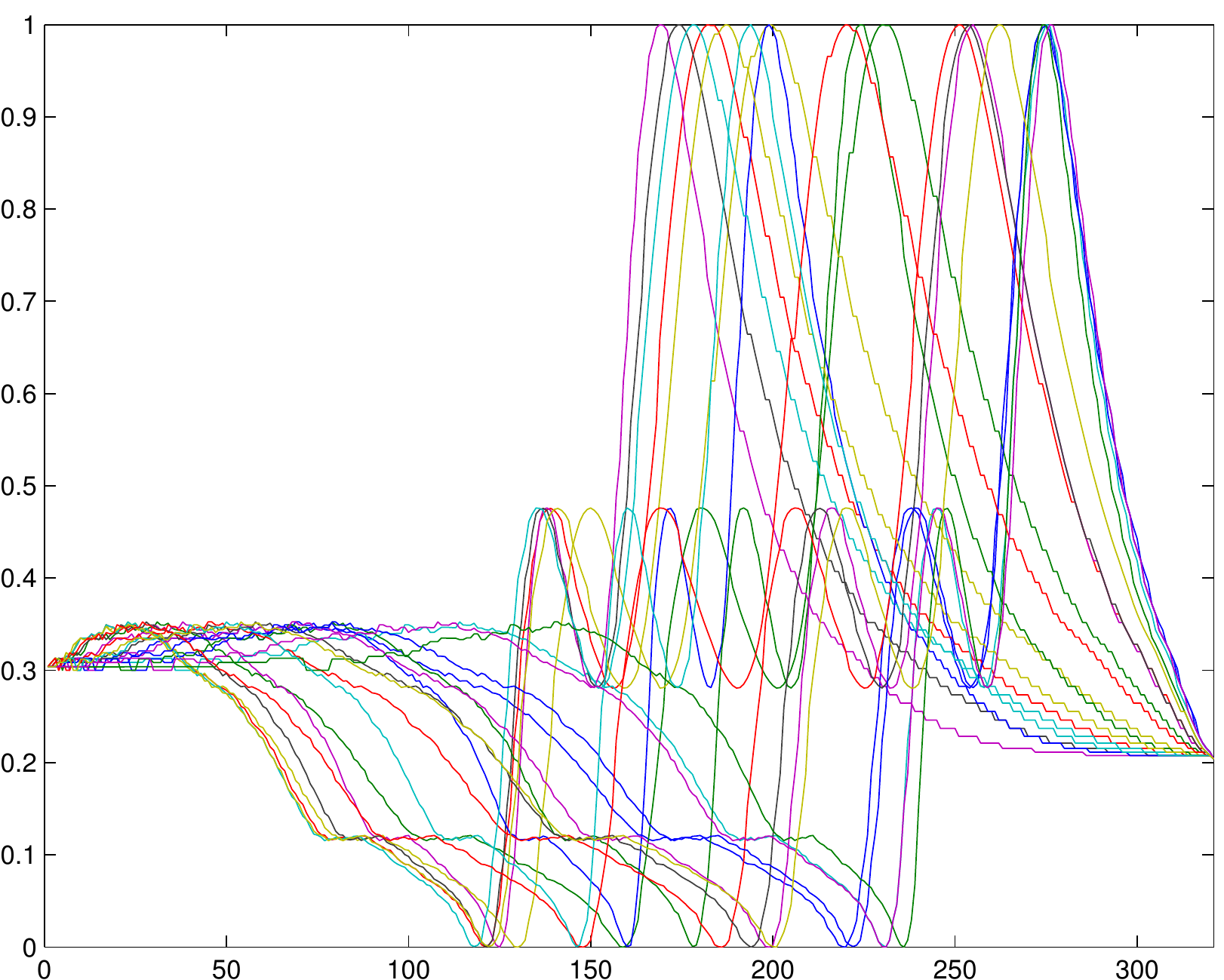}}
	\subfloat[Cluster 3]{
	\includegraphics[width=0.24\linewidth]{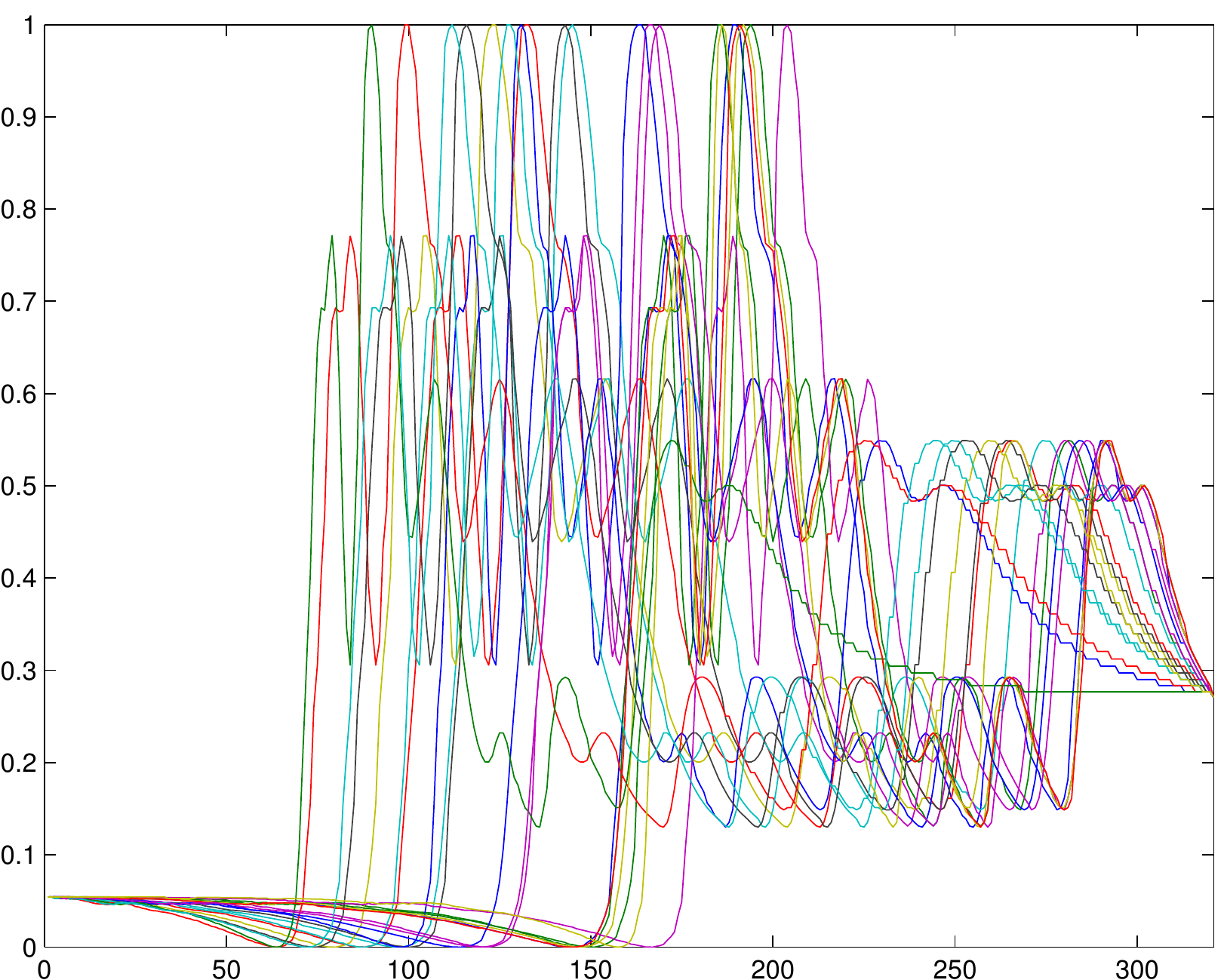}}
	\subfloat[Base Curves]{
	\includegraphics[width=0.24\linewidth]{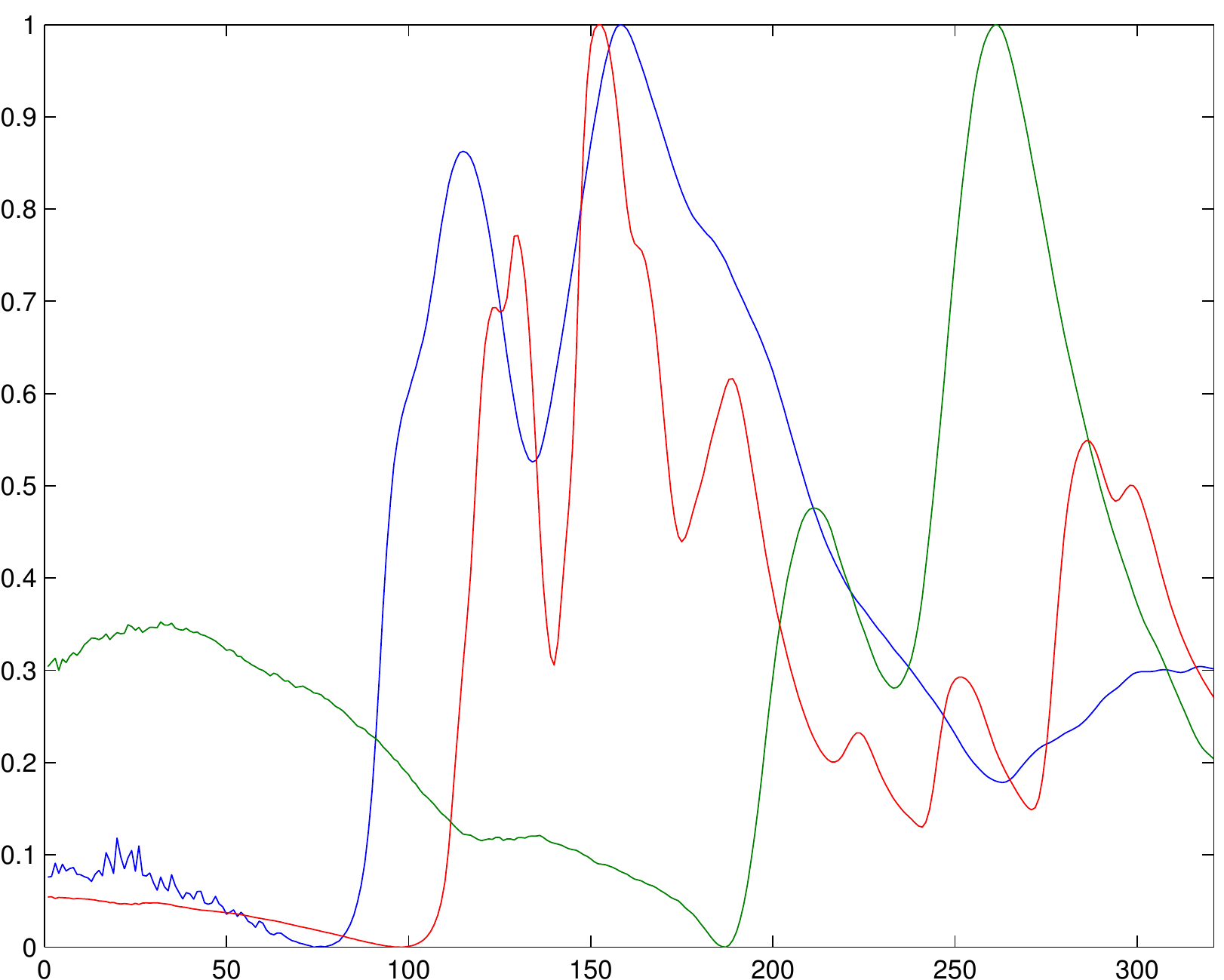}}
	
\caption{Example plots of curves used in the Semi-synthetic TIR Data Experiment. Each cluster has a base curve from the TIR library. The curves for each cluster have been shifted and stretched randomly from the base.}
\label{fig_tir_data}
\end{figure*}

We assemble semi-synthetic data from a library of pure infrared hyper spectral mineral data \citep{guo2013spatial}. For each cluster we pick one spectral sample from the library as a basis. Each curve basis is then randomly shifted and stretched in a random portion. This random warping is performed $20$ times to produce the curves for each cluster. See Figure \ref{fig_tir_data} for an example of data used in this experiment. In this experiment we used three clusters. Again as in the previous experiment we repeated the test $50$ times.

Results are reported in Table \ref{table_tir_results}. The results show that LRR cannot accurately cluster data with this sort of nonlinear invariance, which is commonly found in this type of data due to impurities in the mineral samples. On the other hand cLRR perfectly clustered the data. The closest competitor was the Bayesian method, which also performed well by clustering accurately most of the time. However in some cases the clusters produced were of poor quality, which can be observed in the minimum accuracy and standard deviation statistics. In other words cLRR is far more reliable at clustering this data than the other methods.

\subsection{Handwriting Character Velocities}

\begin{figure} 
\centering
	\subfloat[Trajectories for ``a'']{
	\includegraphics[width=0.25\linewidth]{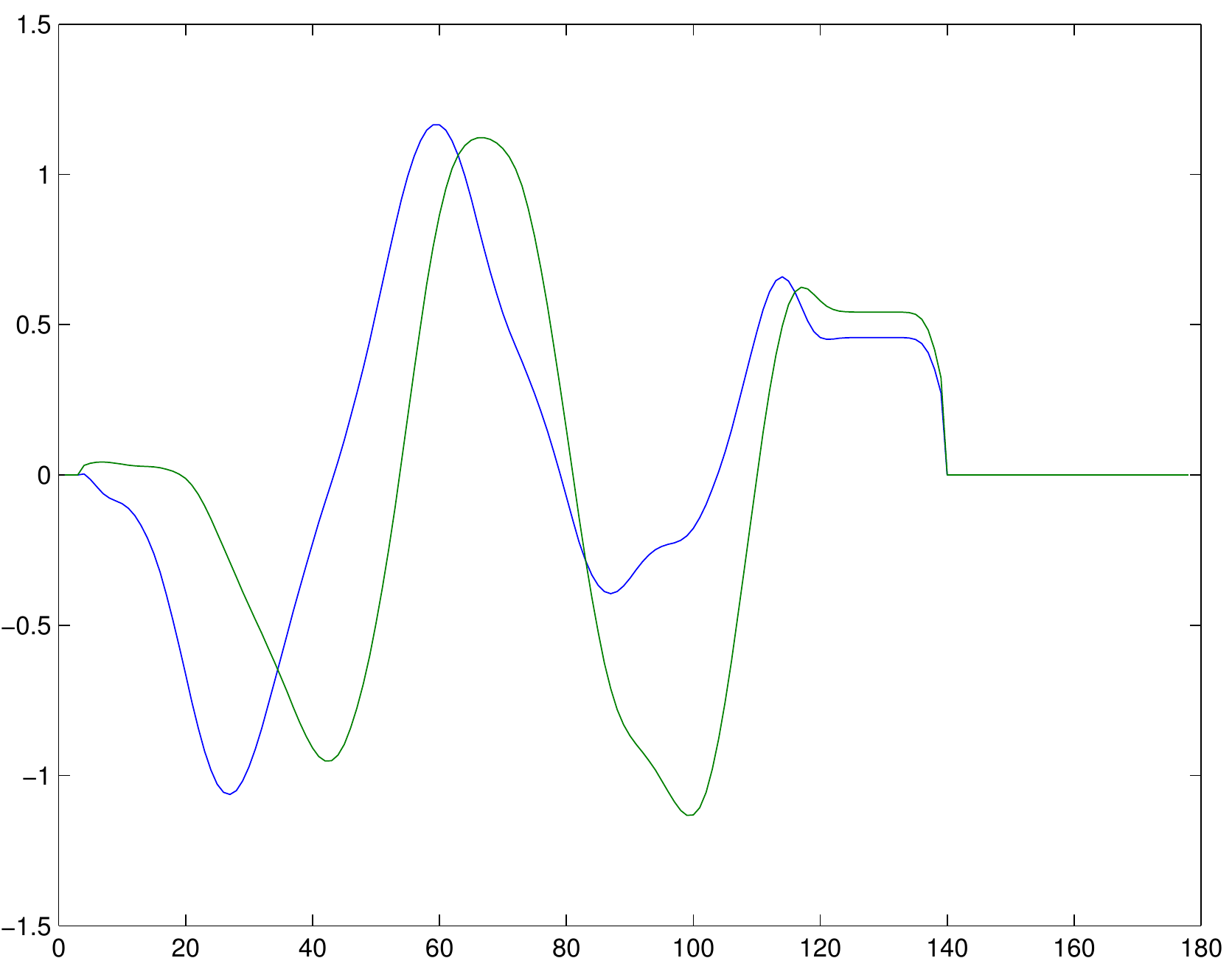}}
	\subfloat[Trajectories for ``b'']{
	\includegraphics[width=0.25\linewidth]{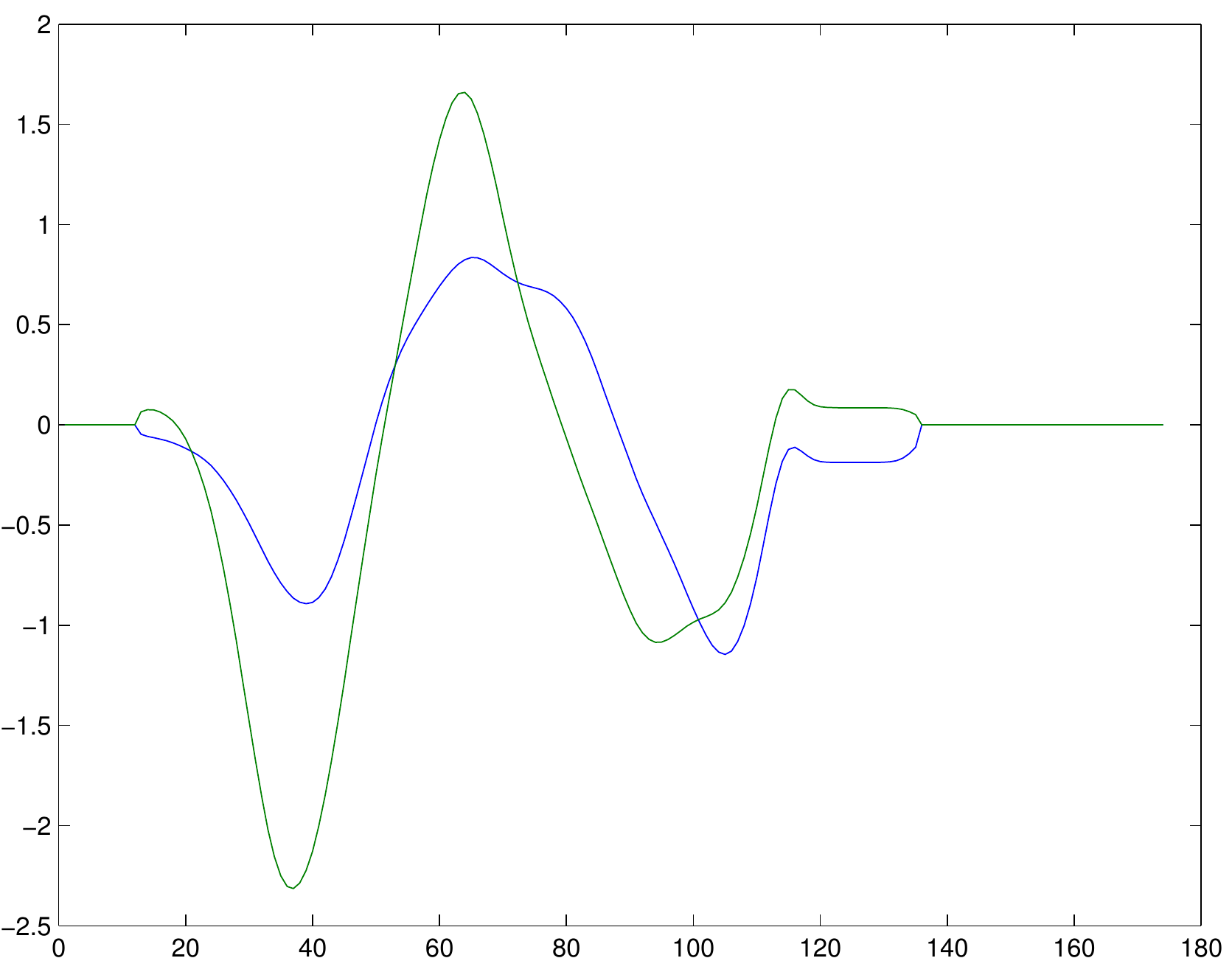}}
	\subfloat[Trajectories for ``c'']{
	\includegraphics[width=0.25\linewidth]{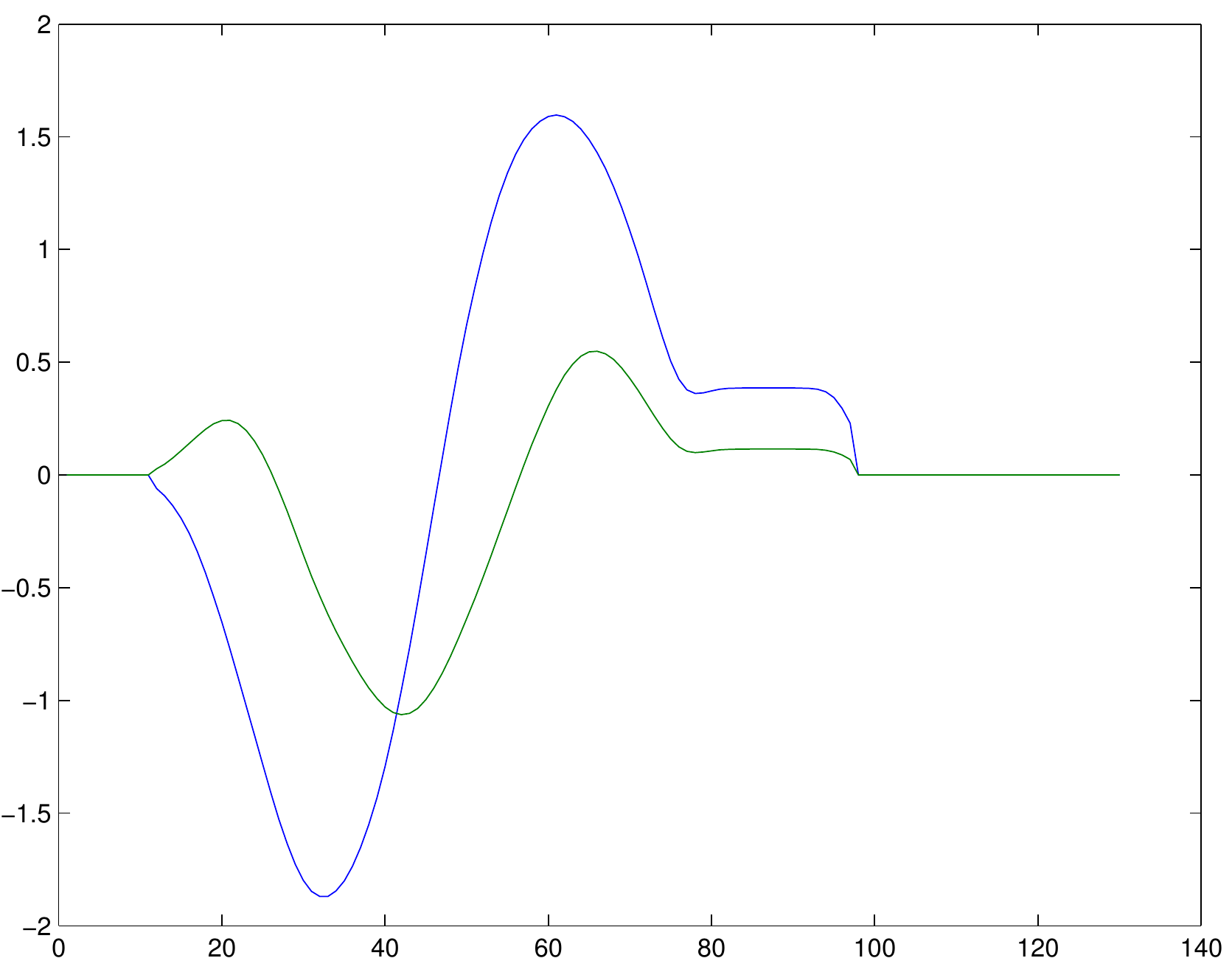}}\\
	\subfloat[Reconstructed ``a'']{
	\includegraphics[width=0.25\linewidth]{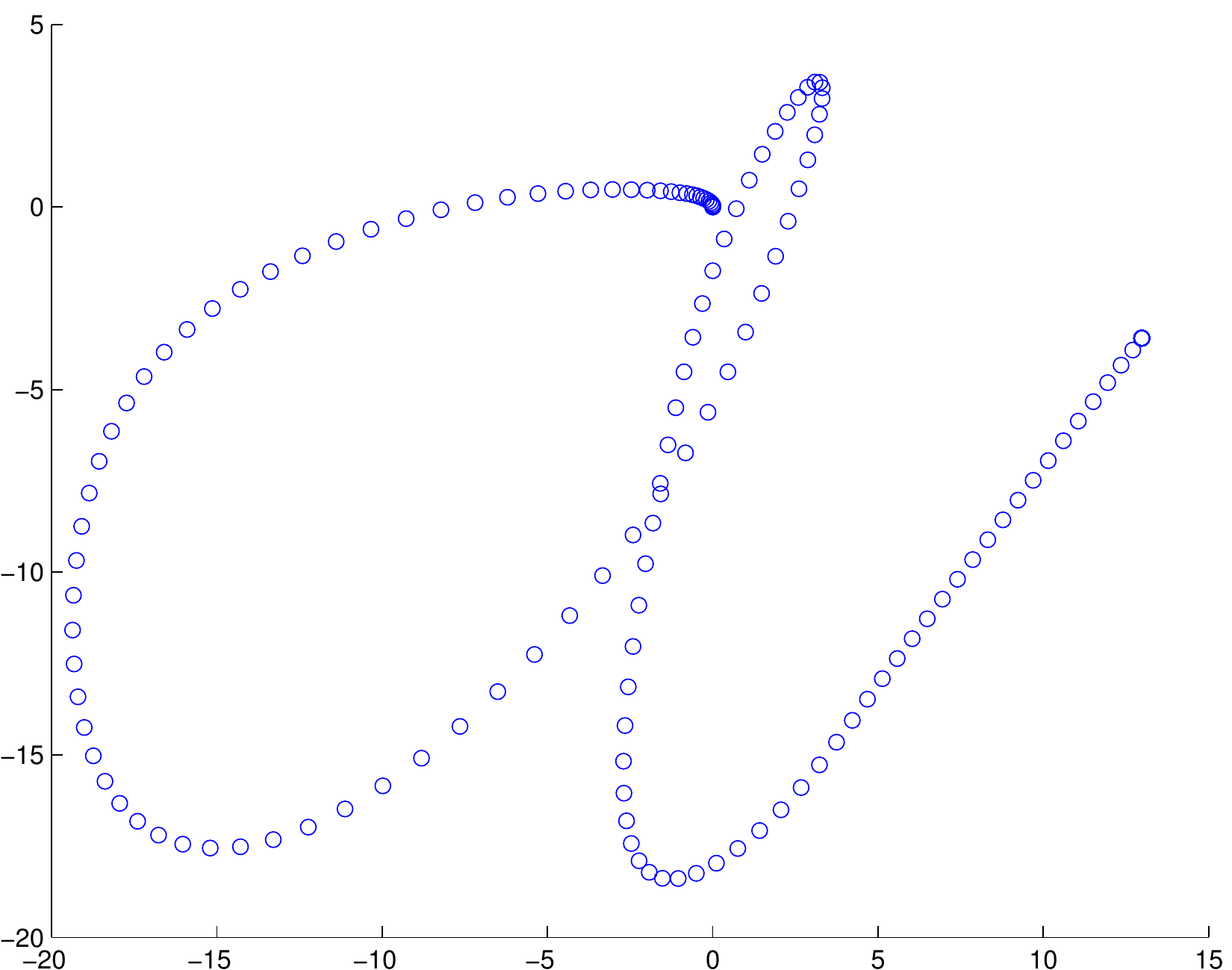}}
	\subfloat[Reconstructed ``b'']{
	\includegraphics[width=0.25\linewidth]{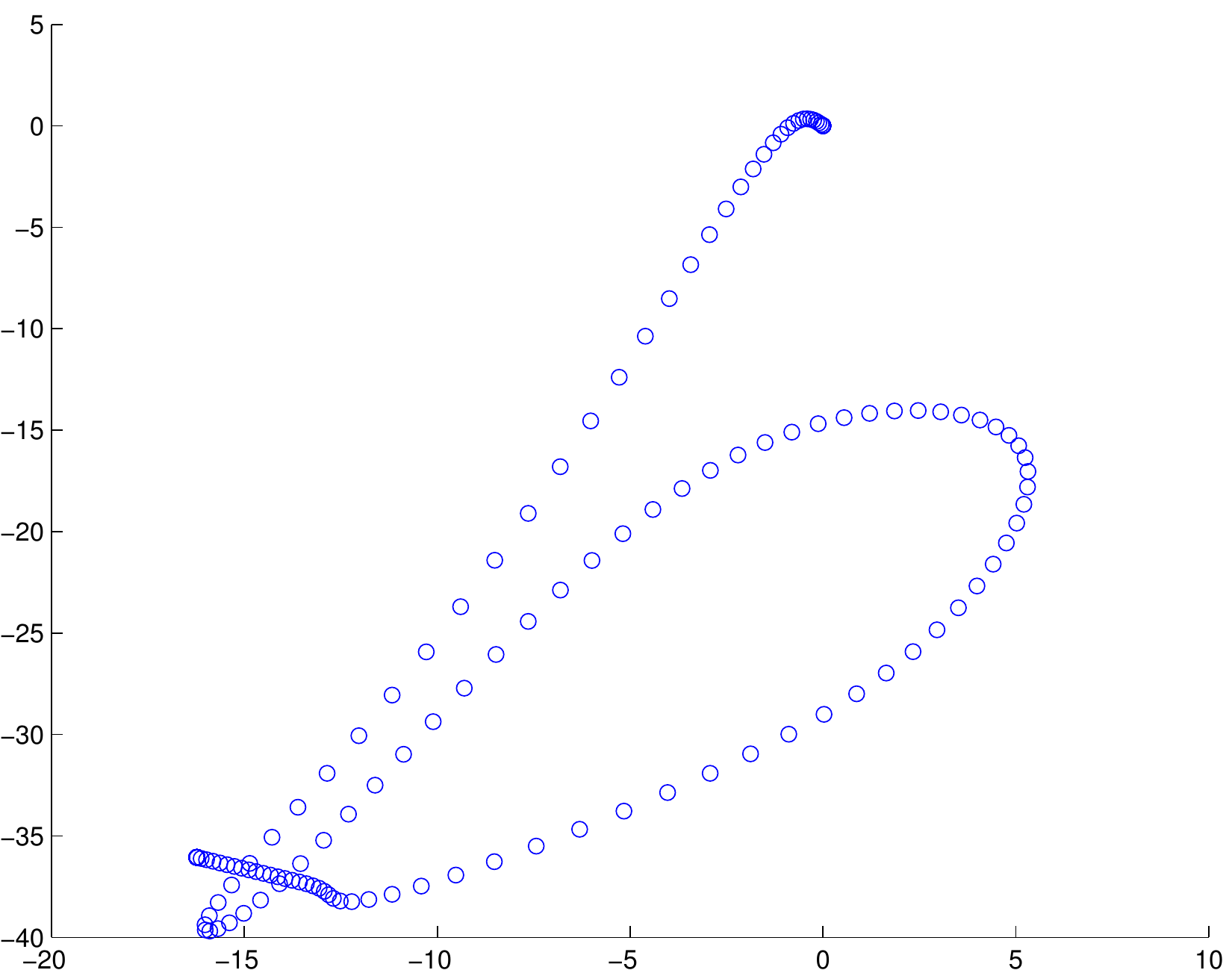}}
	\subfloat[Reconstructed ``c'']{
	\includegraphics[width=0.25\linewidth]{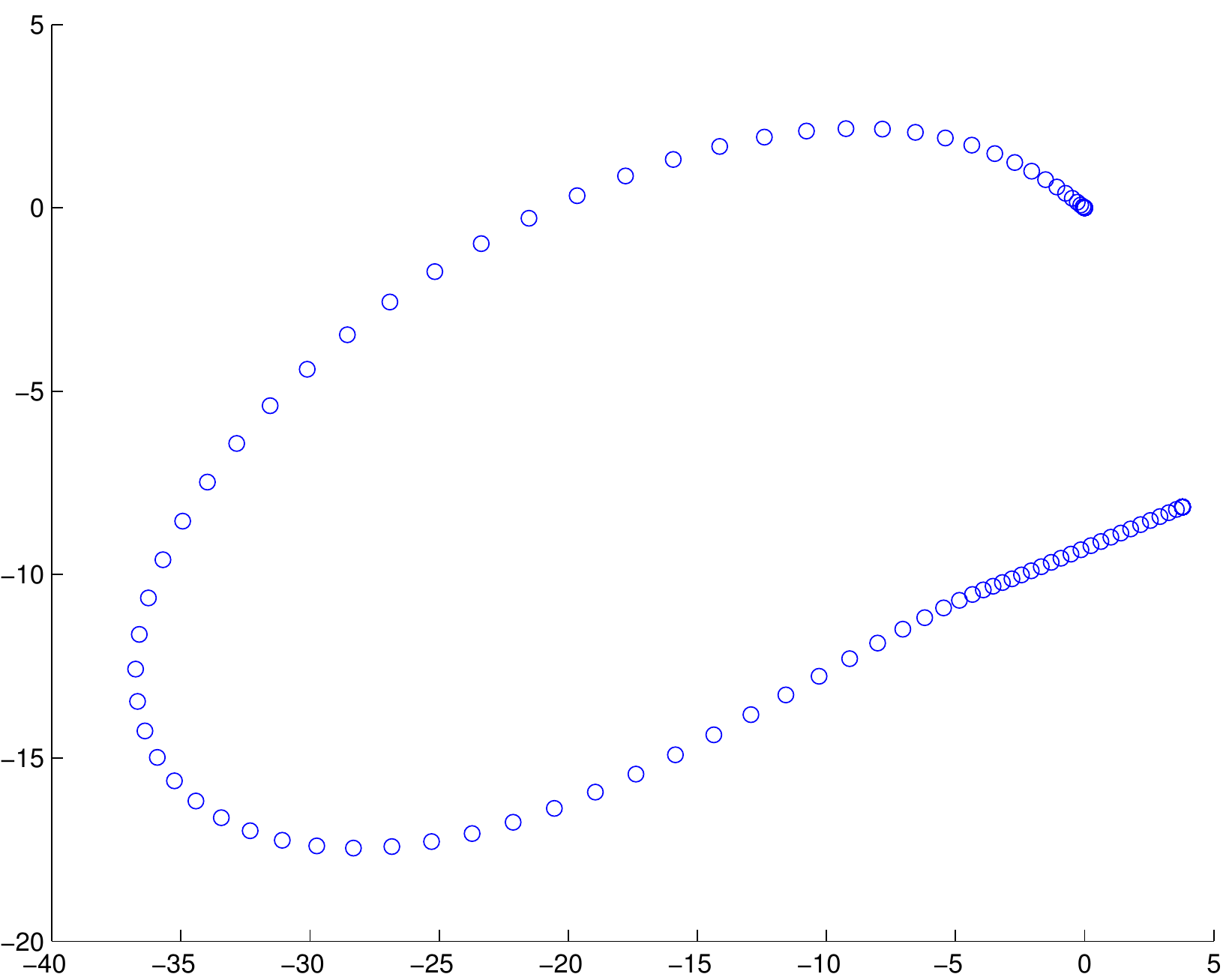}}
\caption{Example data from the character velocity dataset. The top row plots the x and y pen tip velocities over time for three sample characters. The bottom row shows the corresponding character reconstruction by integrating the pen tip velocity data (for visualisation only).}
\label{fig_char_examples}
\end{figure}

\begin{figure}
\centering
	\subfloat[Cluster 1 - X]{
	\includegraphics[width=0.27\linewidth]{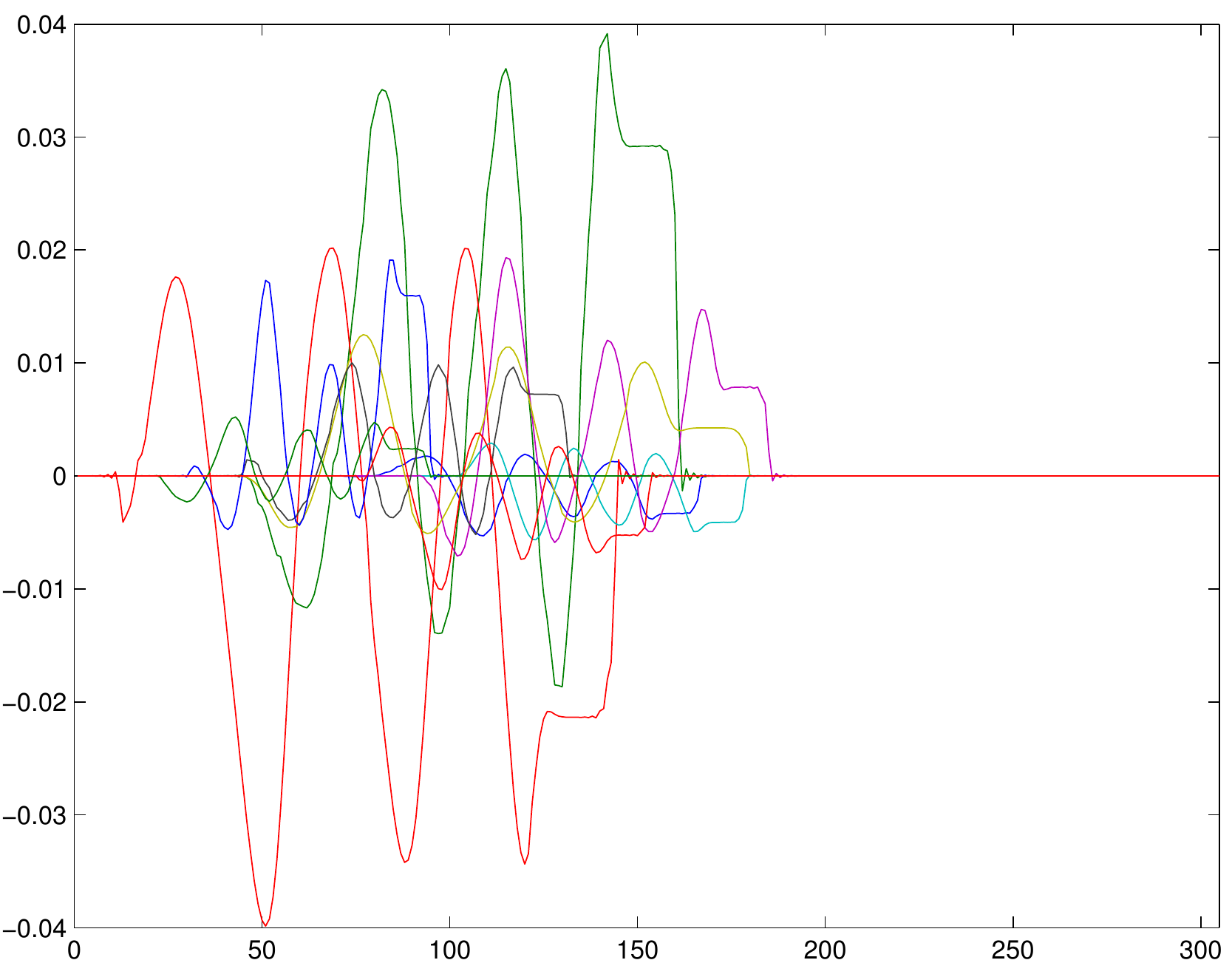}}
	\subfloat[Cluster 2 - X]{
	\includegraphics[width=0.27\linewidth]{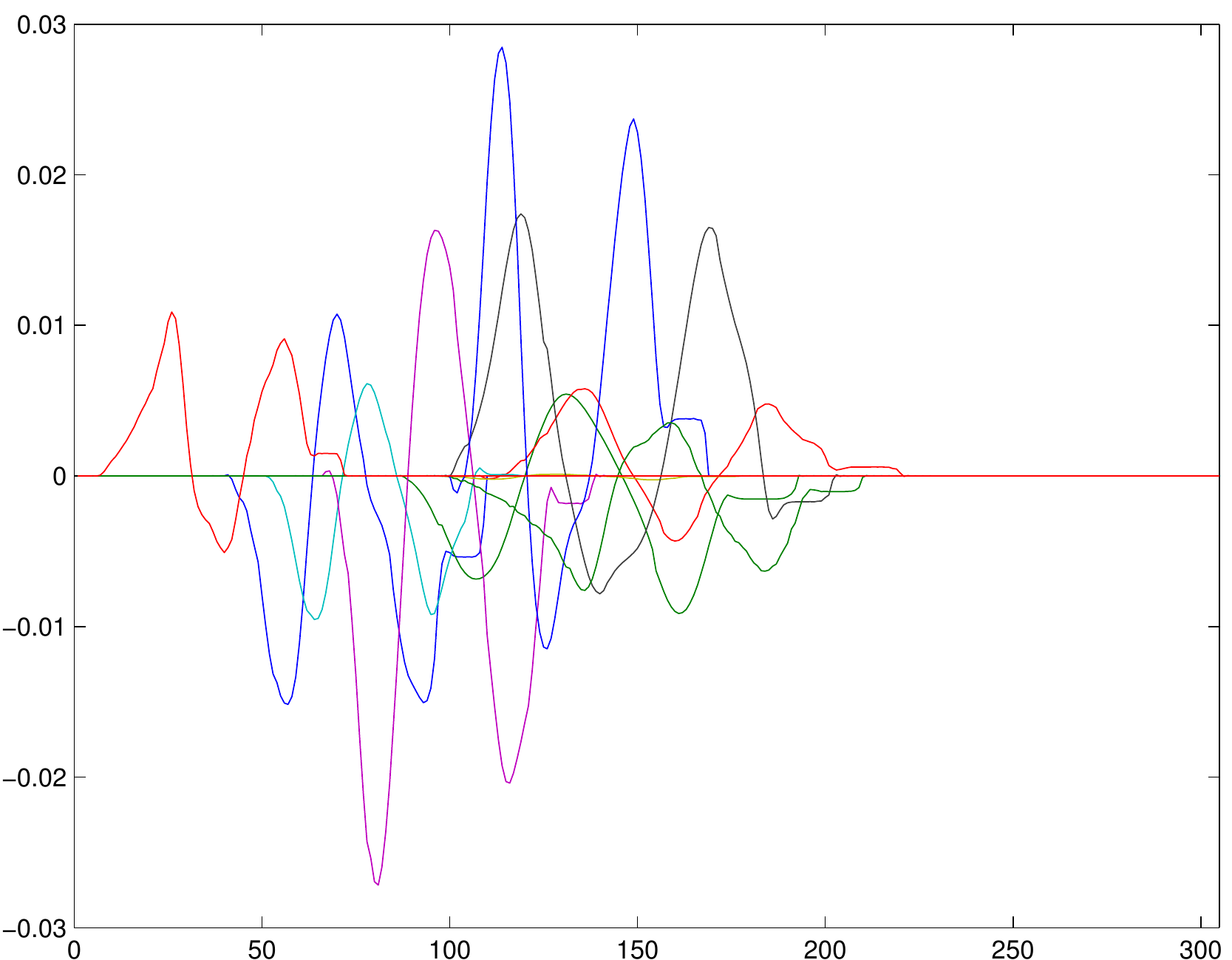}}
	\subfloat[Cluster 3 - X]{
	\includegraphics[width=0.27\linewidth]{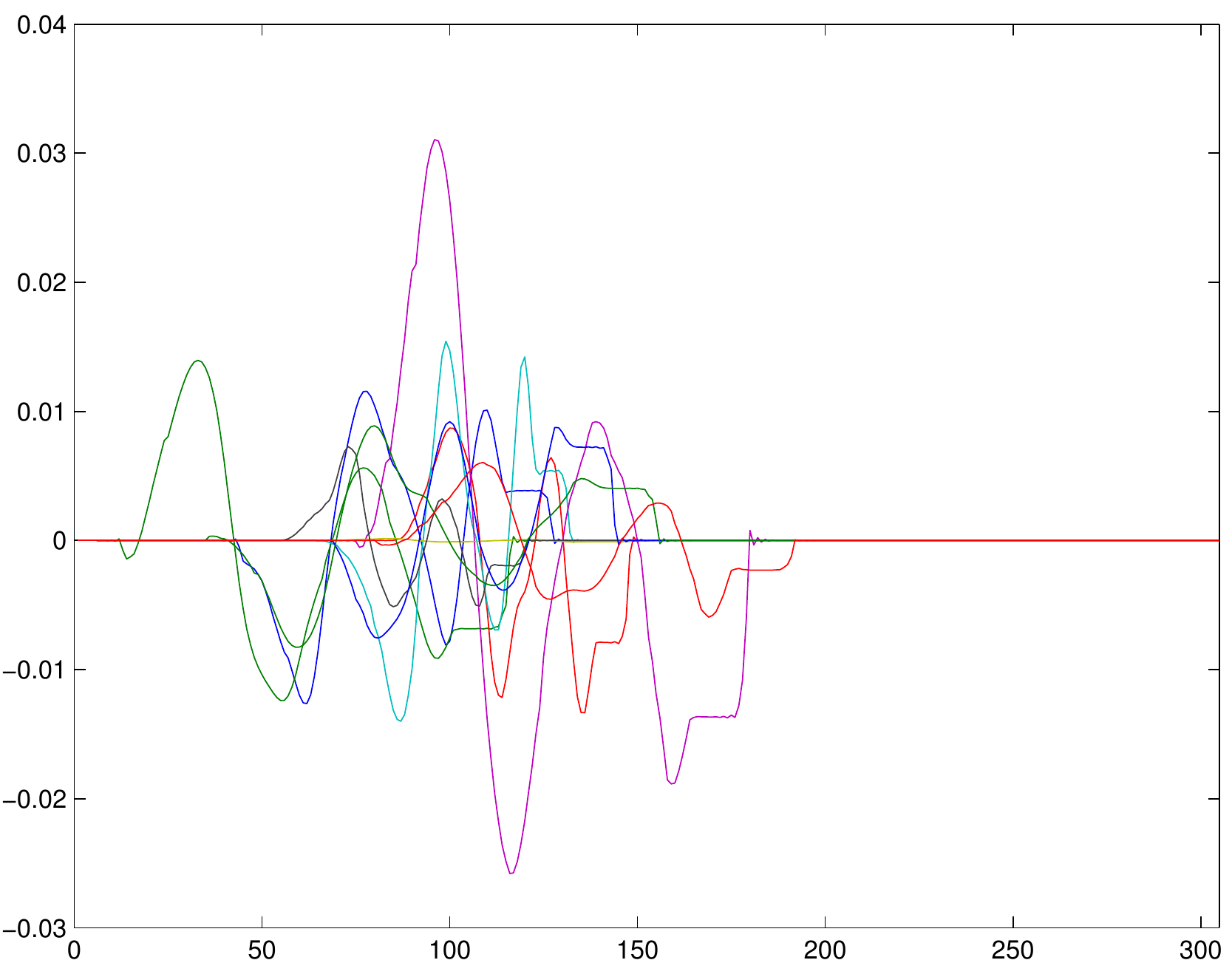}}\\
	\subfloat[Cluster 1 - Y]{
	\includegraphics[width=0.27\linewidth]{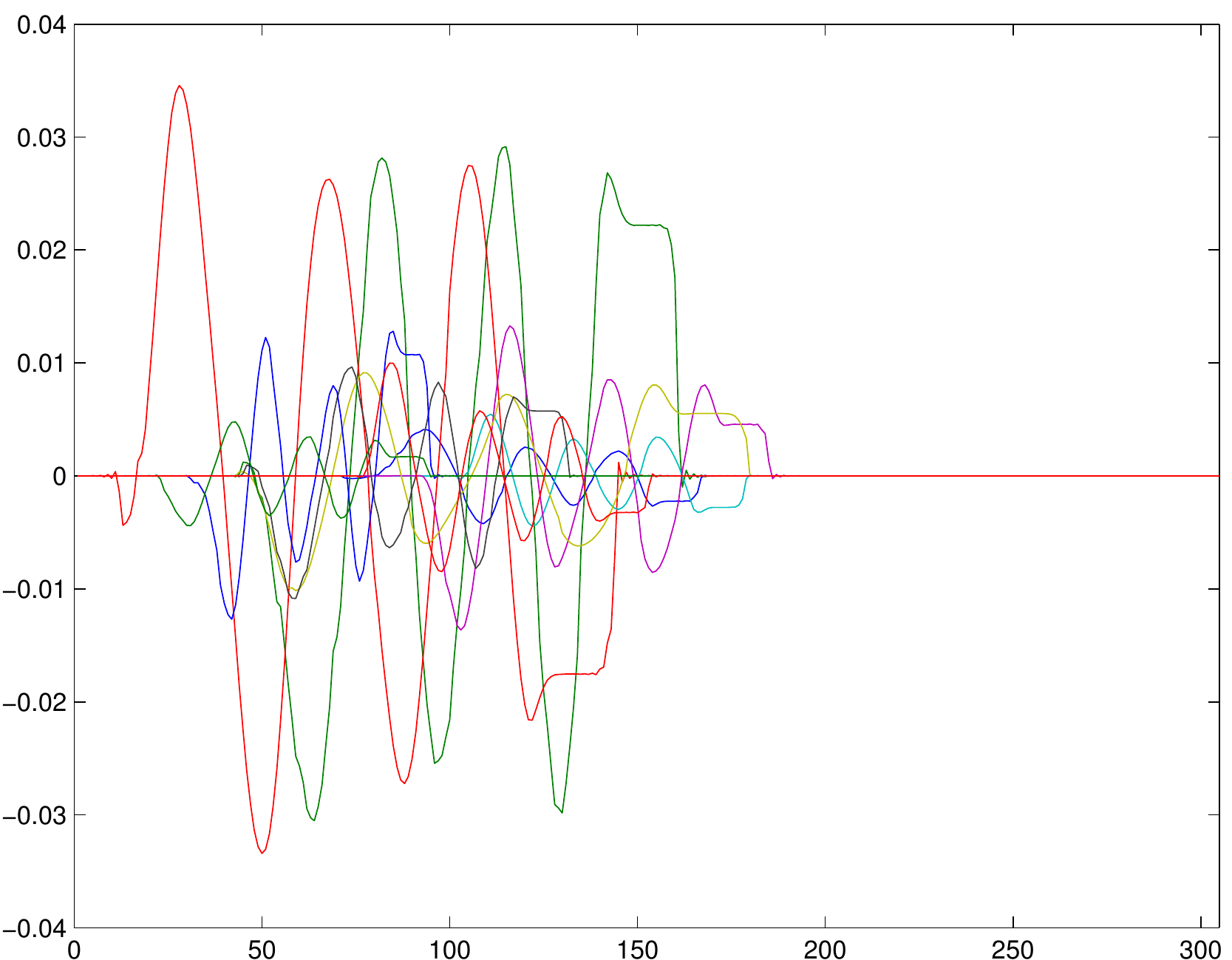}}
	\subfloat[Cluster 2 - Y]{
	\includegraphics[width=0.27\linewidth]{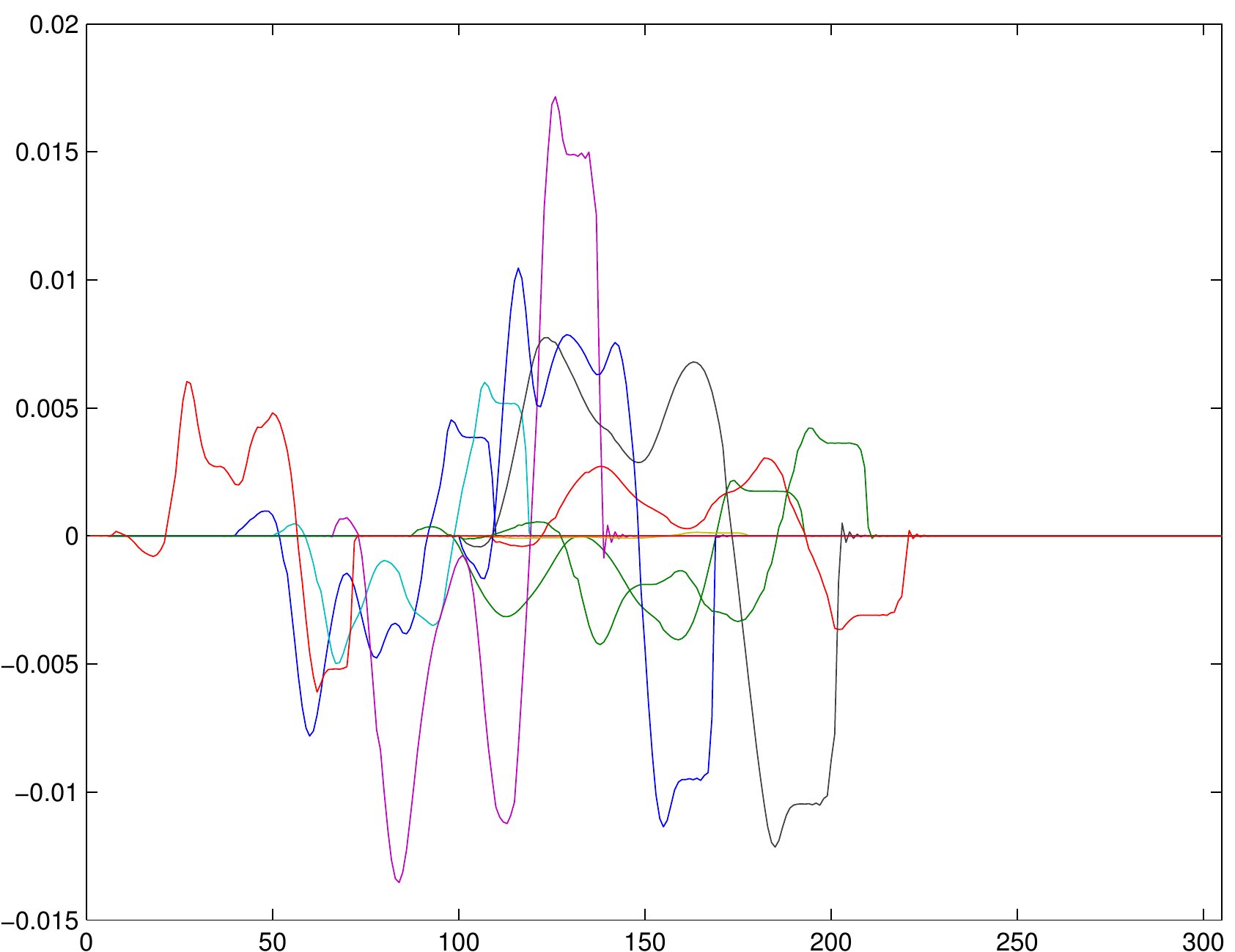}}
	\subfloat[Cluster 3 - Y]{
	\includegraphics[width=0.25\linewidth]{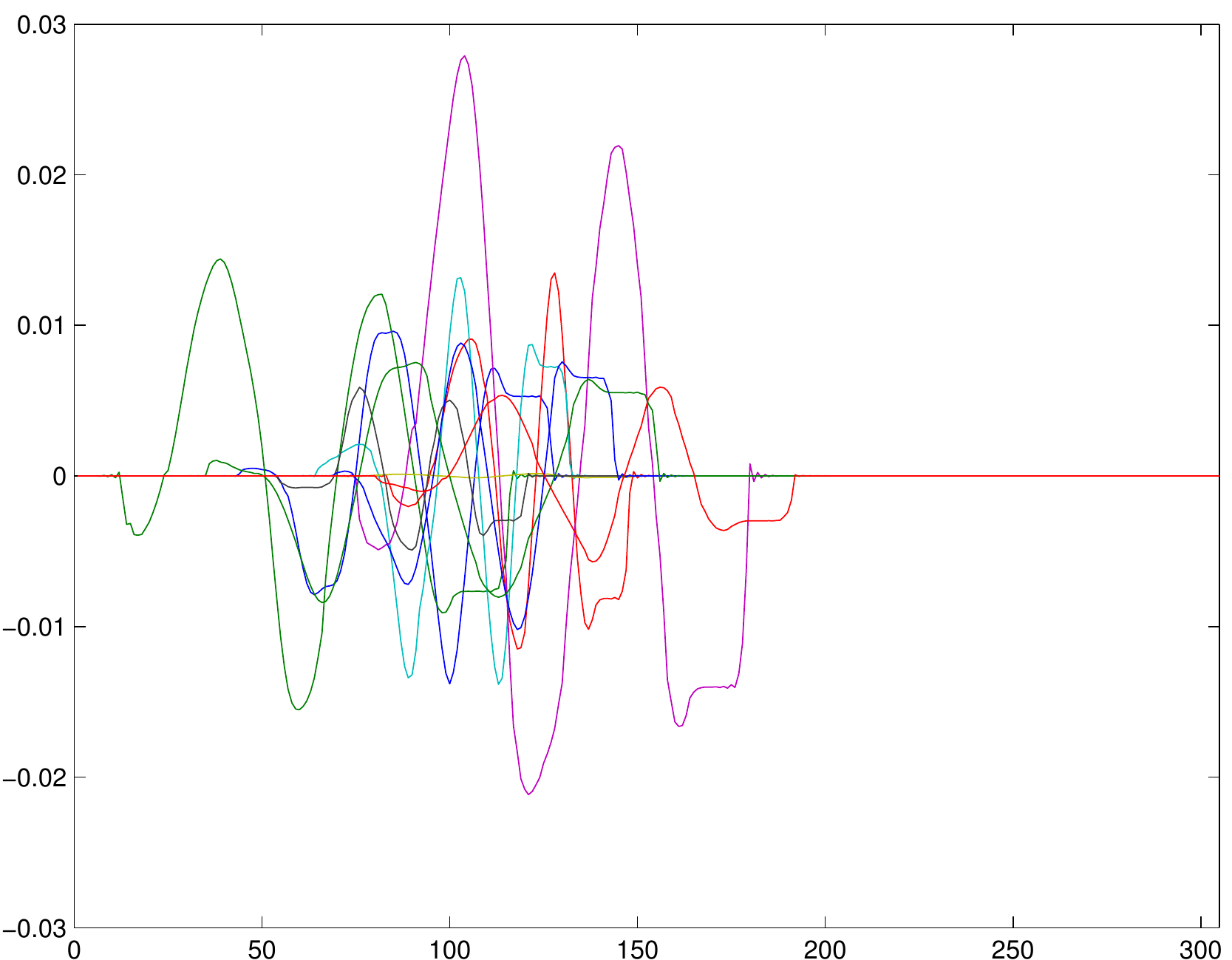}}
\caption{Example plots of curves used in the character velocity experiment. Each cluster consists of randomly selected characters from each class that are then subject to a combination of shifting, warping, stretching or shrinking and scaling. The top row shows the curves from the pen tip velocity in the X direction over time and the bottom row shows the same but for the Y direction.}
\label{fig_char_data}
\end{figure}

\begin{table*}
\centering
\begin{tabular}{c c c c c c c}
\hline
 & Mean & Median & Max & Min & Std & Mean Run Time (s) \\
\hline

kmeans & 0.0\% & 0.0\% & 0.0\% & 0.0\% & 0.0\% & 0.00 \\ 
DTW & 48.2\% & 46.7\% & 63.3\% & 40.0\% & 6.4\% & 0.18 \\ 
LRR & 48.3\% & 46.7\% & 70.0\% & 36.7\% & 7.7\% & 0.01 \\ 
Bayesian & 47.5\% & 33.3\% & 83.3\% & 33.3\% & 16.3\% & 79.19 \\ 
CurveLRR & \hl{84.6\%} & \hl{86.7\%} & \hl{100.0\%} & \hl{53.3\%} & 14.4\% & 46.38 \\ 

\end{tabular}

\caption{Character Velocity Results}
\label{table_char_results}
\end{table*}

In this experiment a real world dataset consisting of a collection of the pen tip trajectories of handwritten English characters were used to evaluate performance. The dataset consists of pen position data collected by a digitisation tablet at 200Hz, which is then converted to horizontal and vertical velocities \citep{williams2006extracting, williams2008modelling}. These 2-D trajectory curves are normalised such that the mean of each curve is close to zero. See Figure \ref{fig_char_examples} for some examples of this data. Figure \ref{fig_char_data} shows the example plots of curves used in the character classification experiment.

For each run of this test, twenty characters were randomly selected from three random character classes. The data as originally released has been carefully produced and processed so that trajectories for each characters are extremely similar. Far more so than is realistic. For example the start time for each character has been aligned furthermore the writing speed, character size and variance in velocity over time is extremely consistent. Therefore to make the data more realistic we randomly globally shift each character so that their start times vary. Furthermore we randomly globally stretch and shrink each trajectory to account for different writing speeds, we also scale the trajectories by applying constant factors to account for character size and we lastly perform local warping (as done in the semi-synthetic experiment) to account for variance in speed over time. 

Results can be found in Table \ref{table_char_results}. cLRR shows excellent performance with a median accuracy of $80\%$ on an extremely challenging dataset. The closest competitors only reach a median clustering accuracy of $50\%$, which is extremely poor. It is clear to see that cLRR clearly outperforms the other methods in all metrics. 

\subsection{Handwriting Character Trajectories}

\begin{figure}
\centering
	\subfloat[``u'' character]{
	\includegraphics[width=0.1\linewidth]{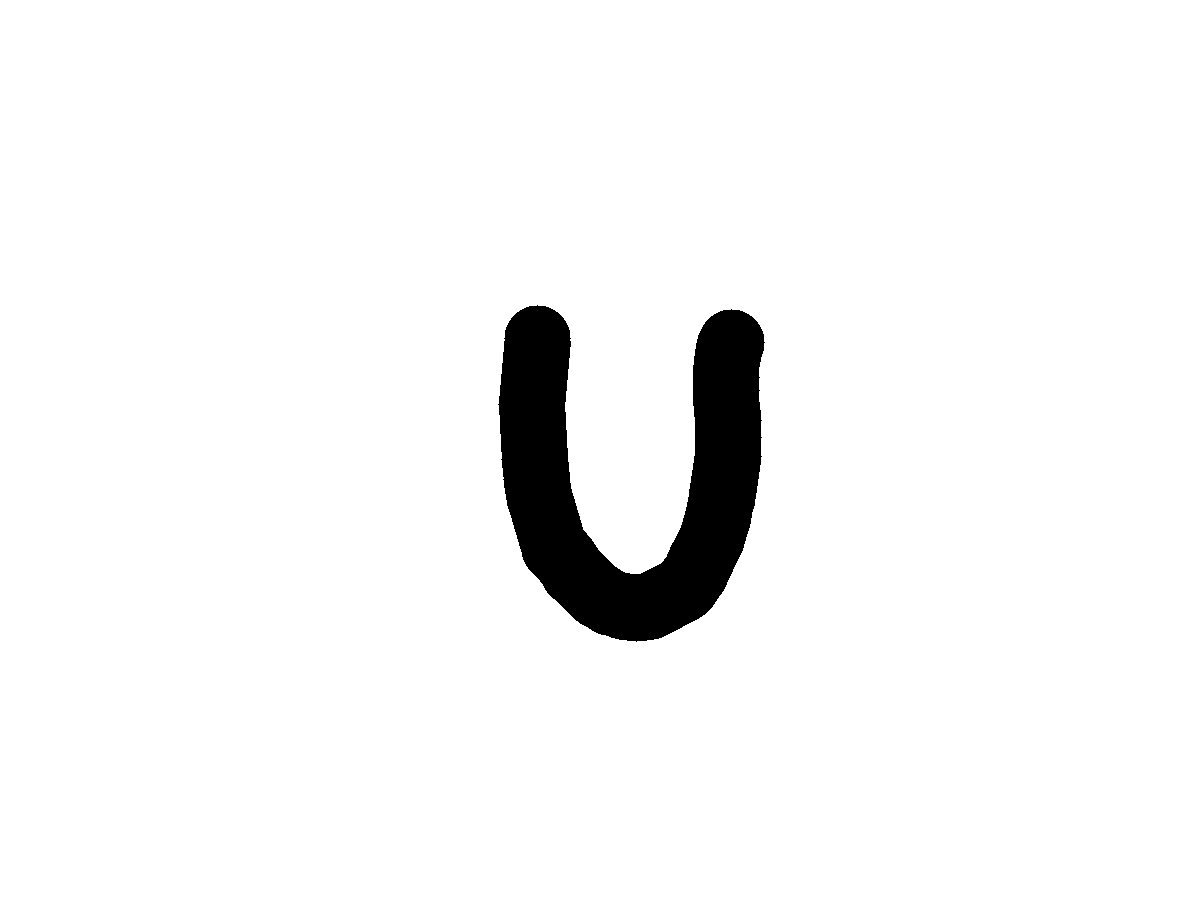}
	\includegraphics[width=0.1\linewidth]{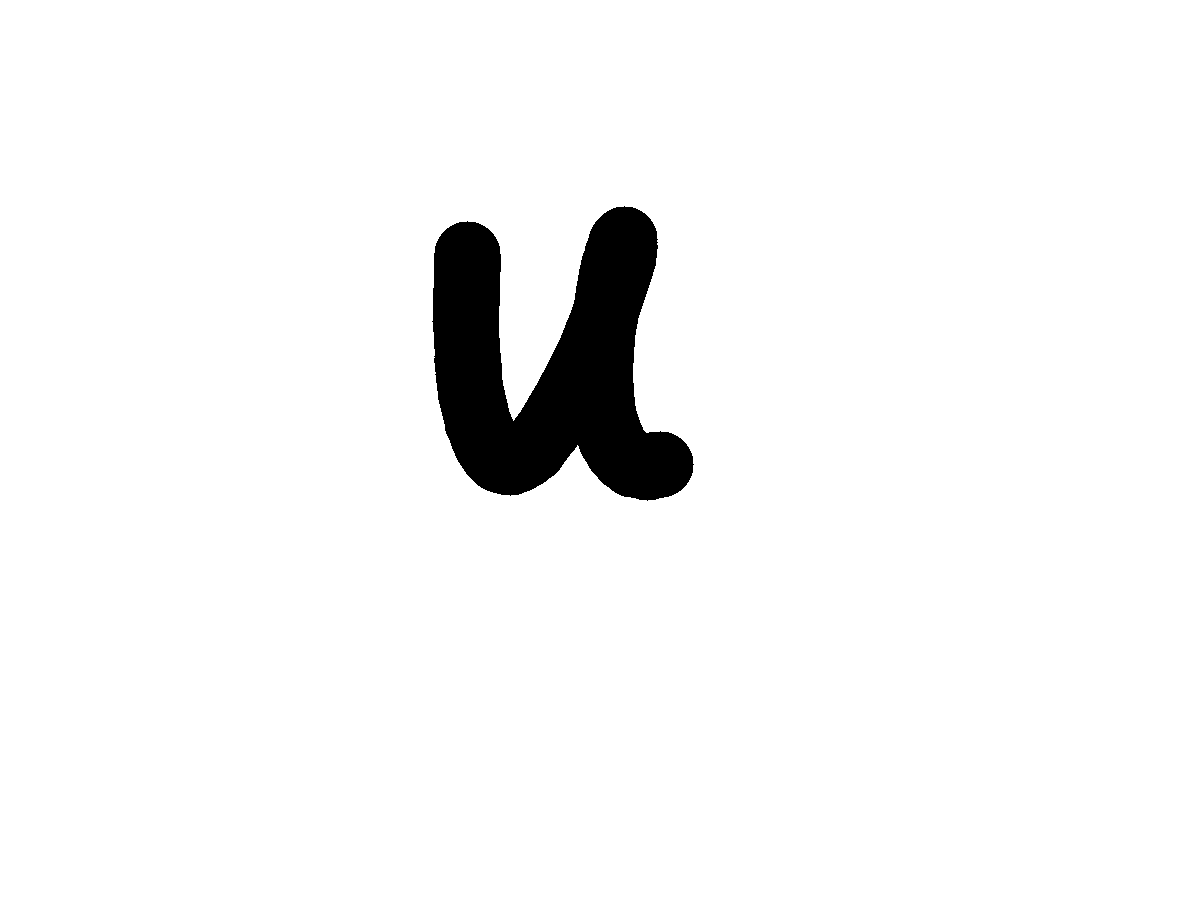}
	\includegraphics[width=0.1\linewidth]{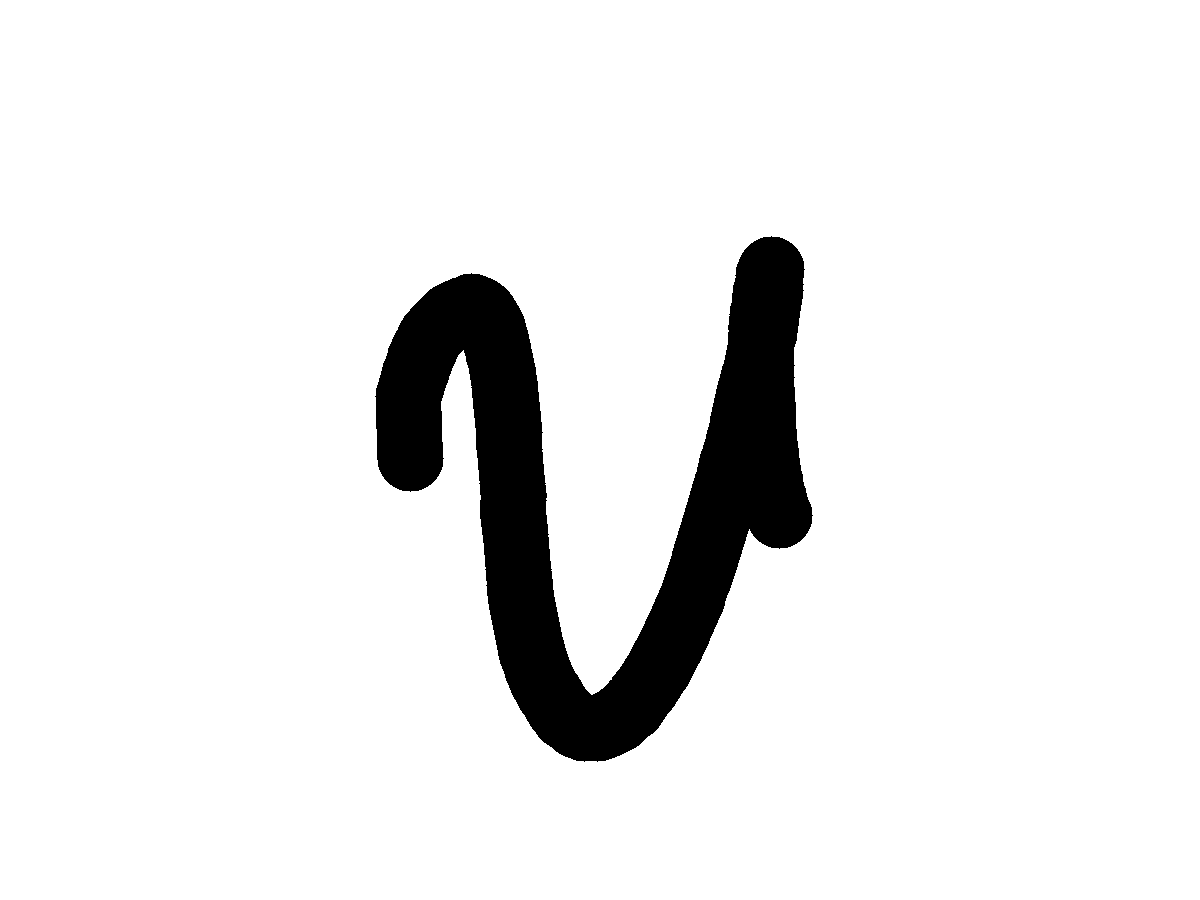}}
	\subfloat[``1'' character]{
	\includegraphics[width=0.1\linewidth]{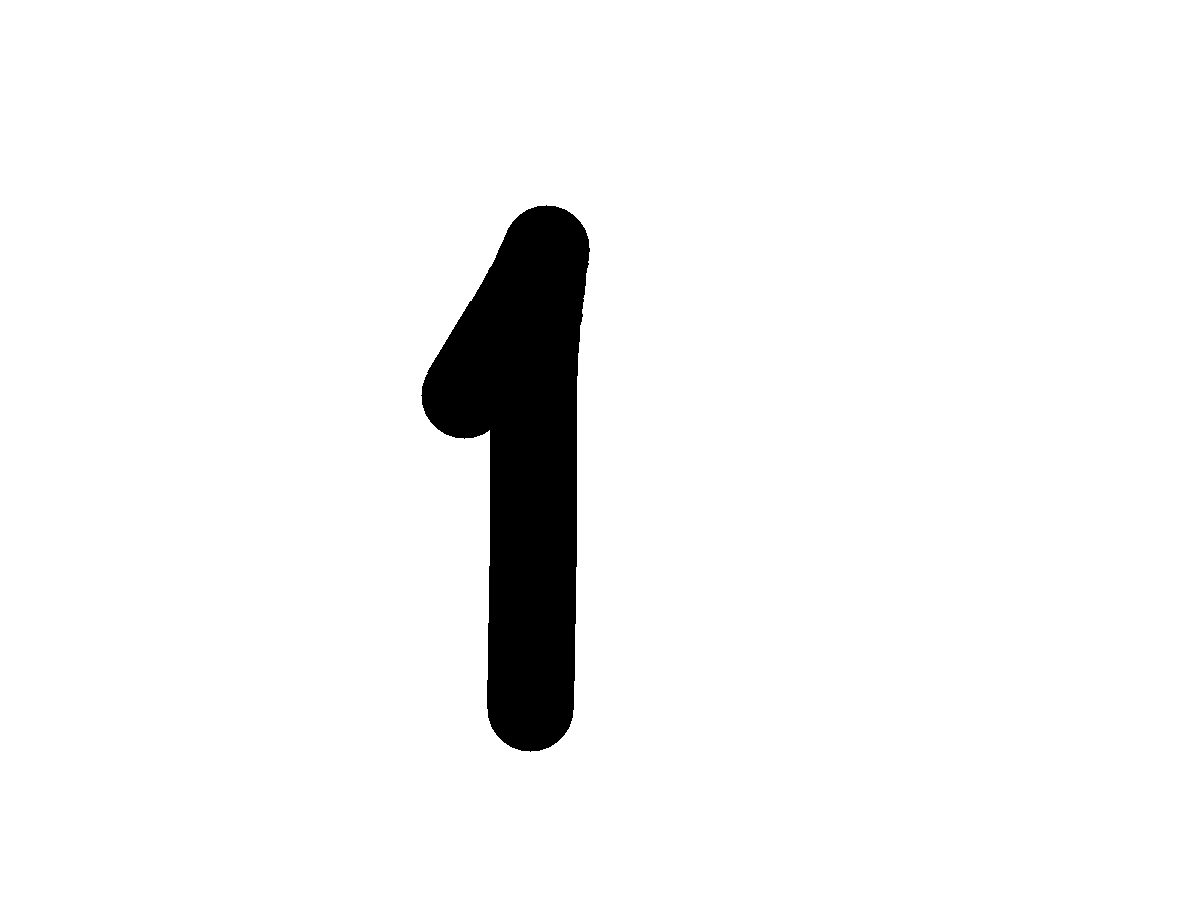}
	\includegraphics[width=0.1\linewidth]{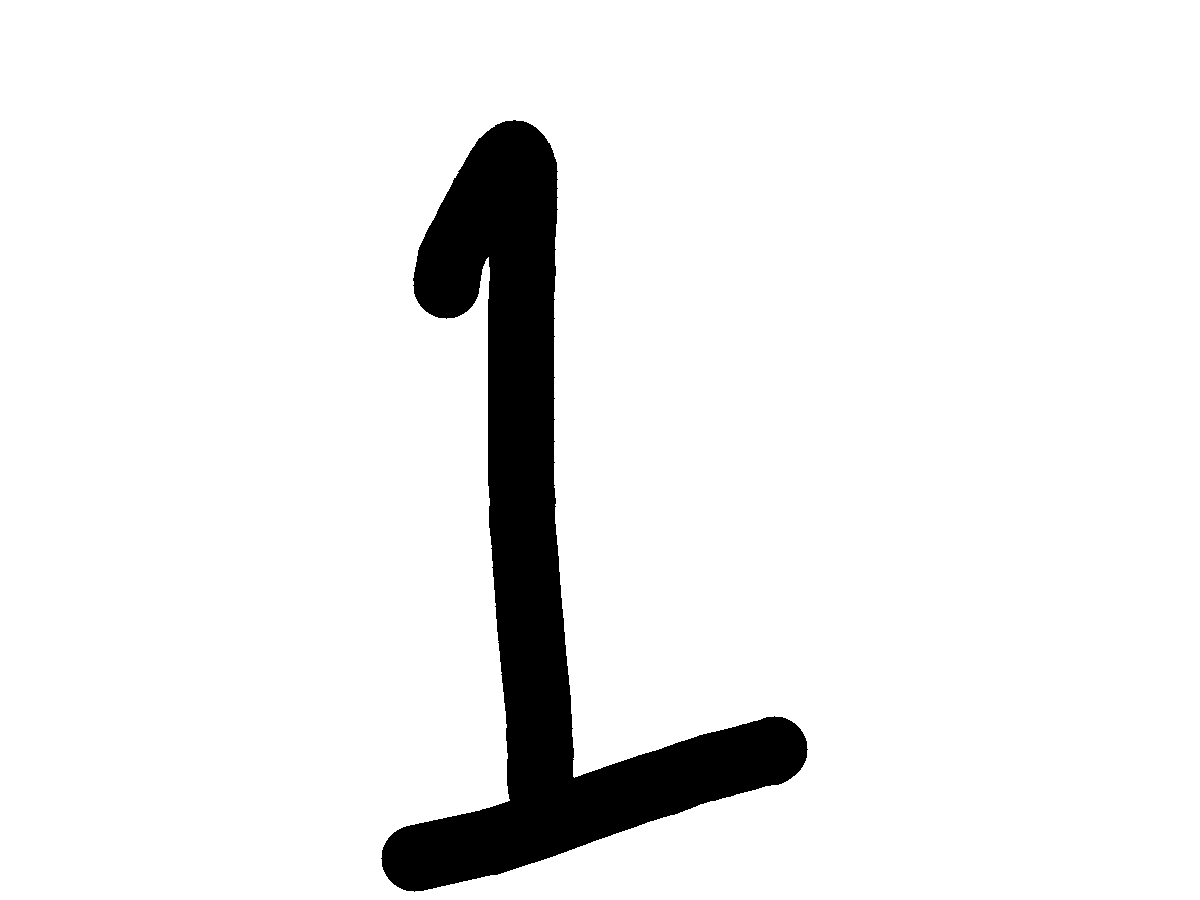}
	\includegraphics[width=0.1\linewidth]{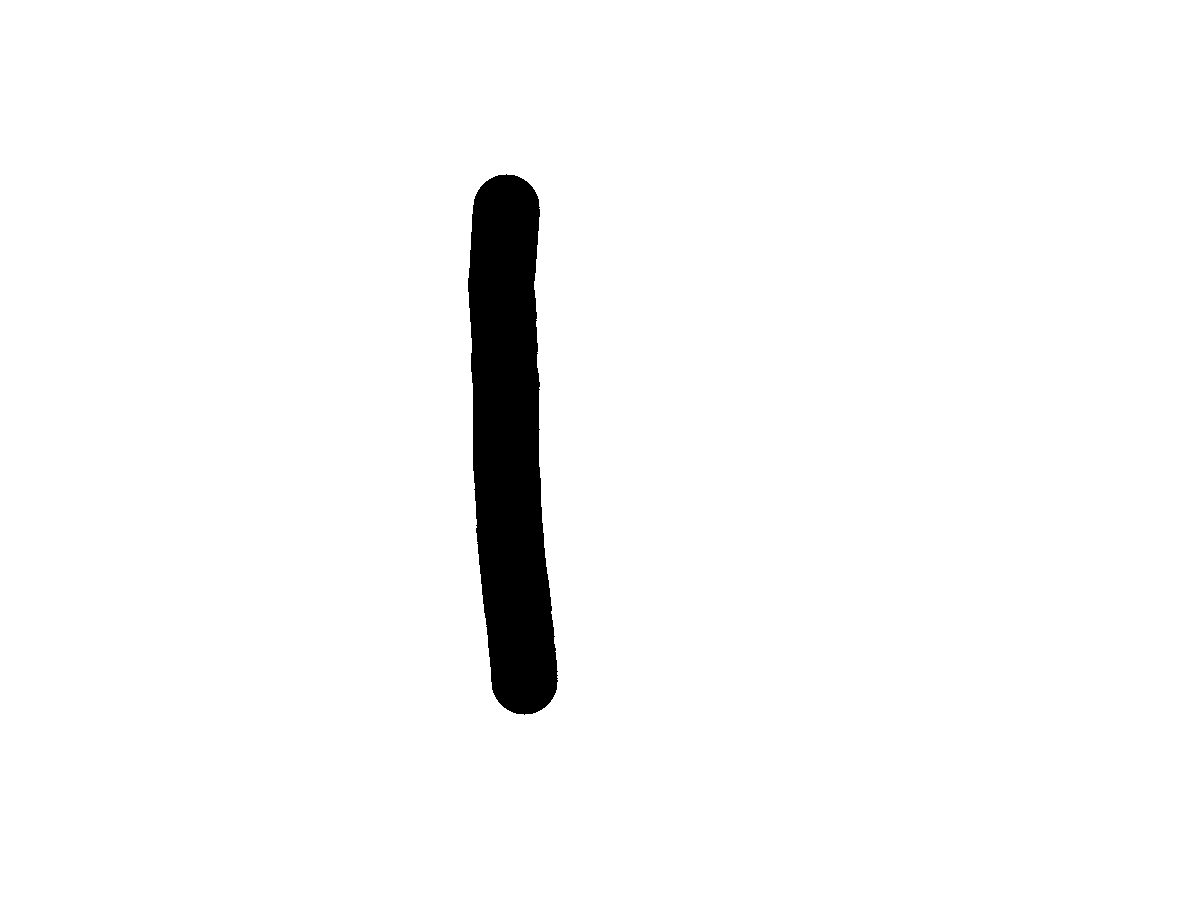}}\\
	\subfloat[``4'' character]{
	\includegraphics[width=0.1\linewidth]{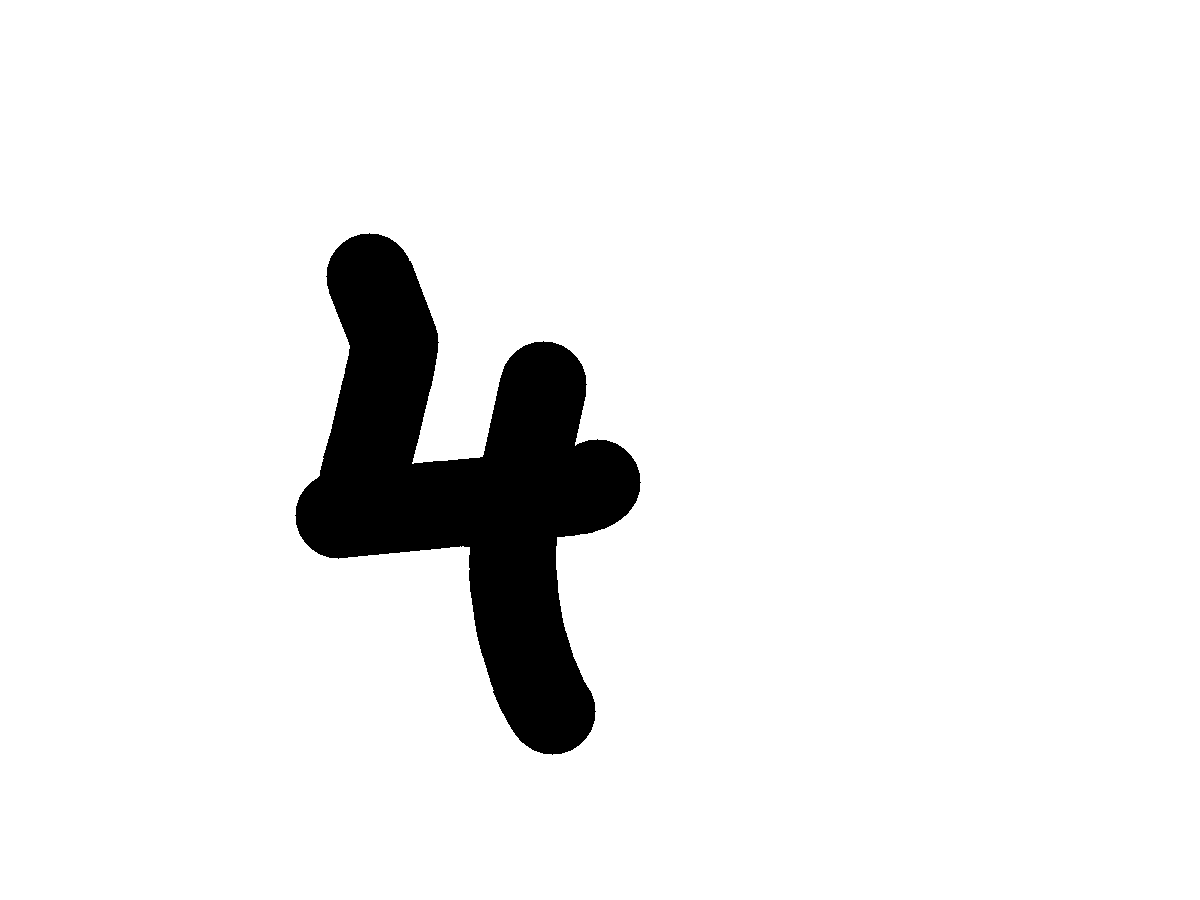}
	\includegraphics[width=0.1\linewidth]{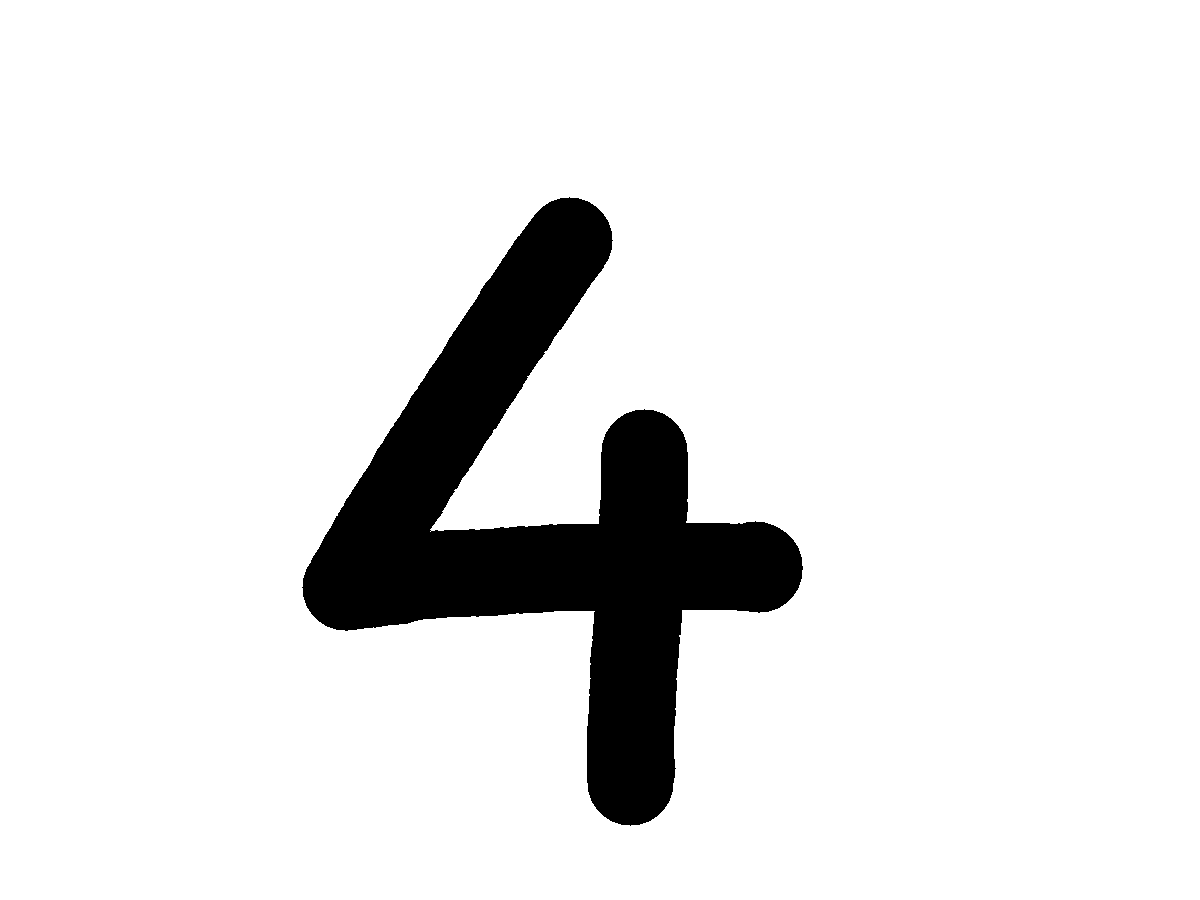}
	\includegraphics[width=0.1\linewidth]{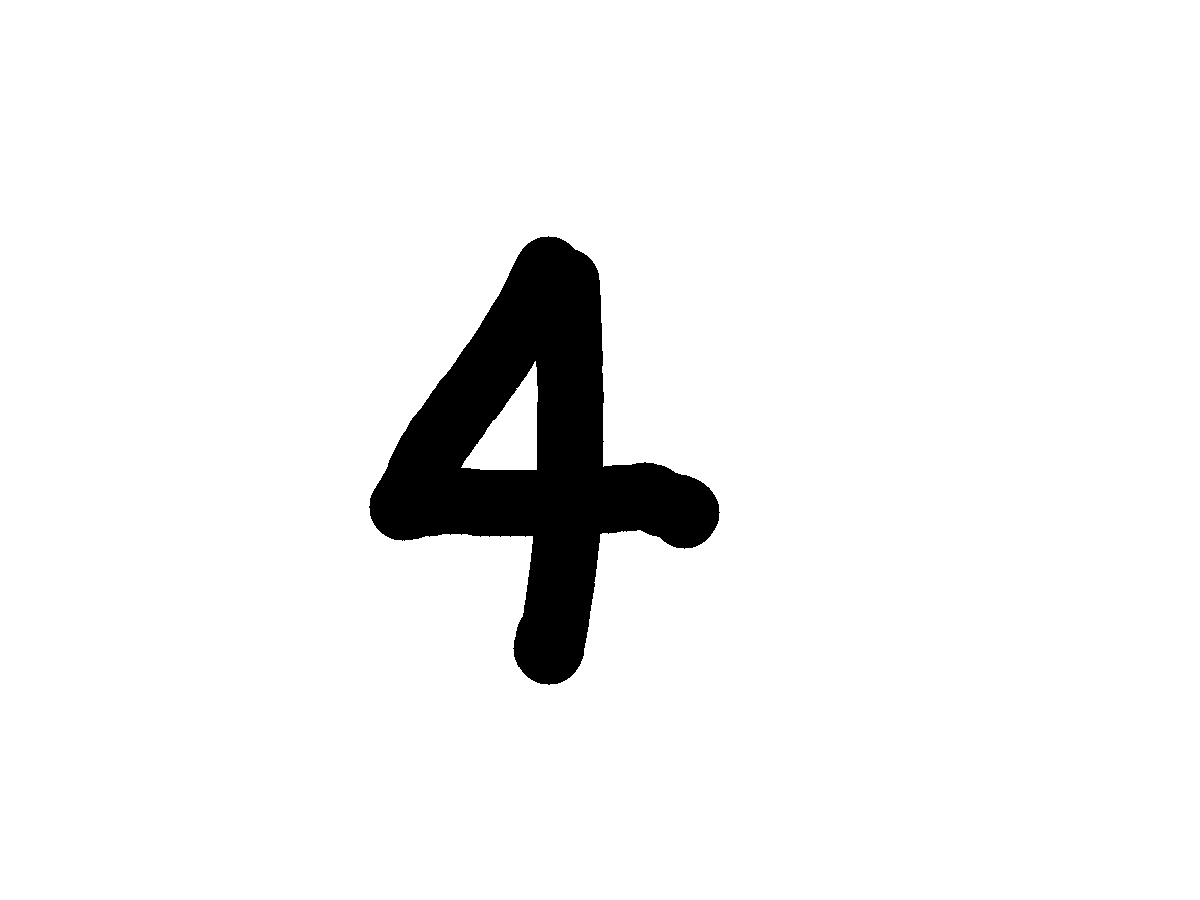}}
		\subfloat[``n'' character]{
	\includegraphics[width=0.1\linewidth]{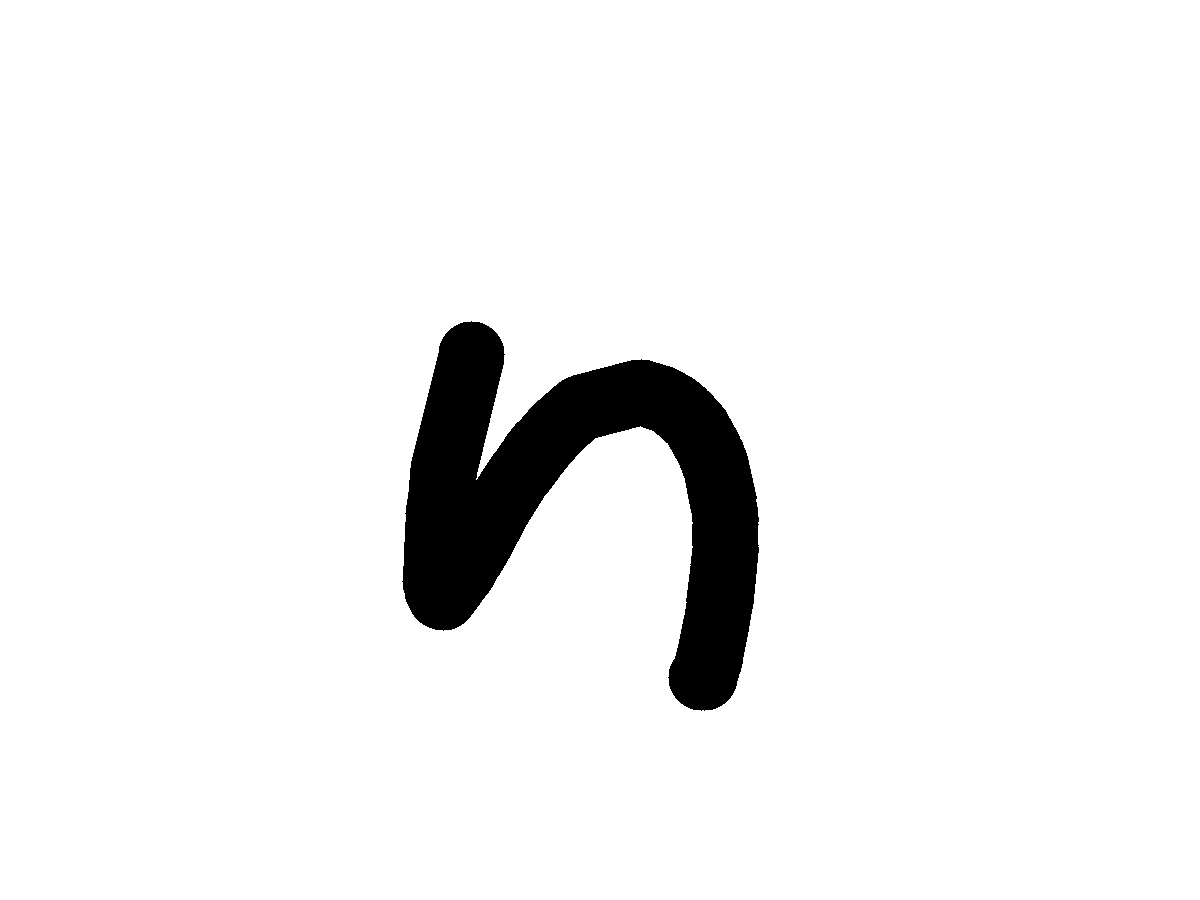}
	\includegraphics[width=0.1\linewidth]{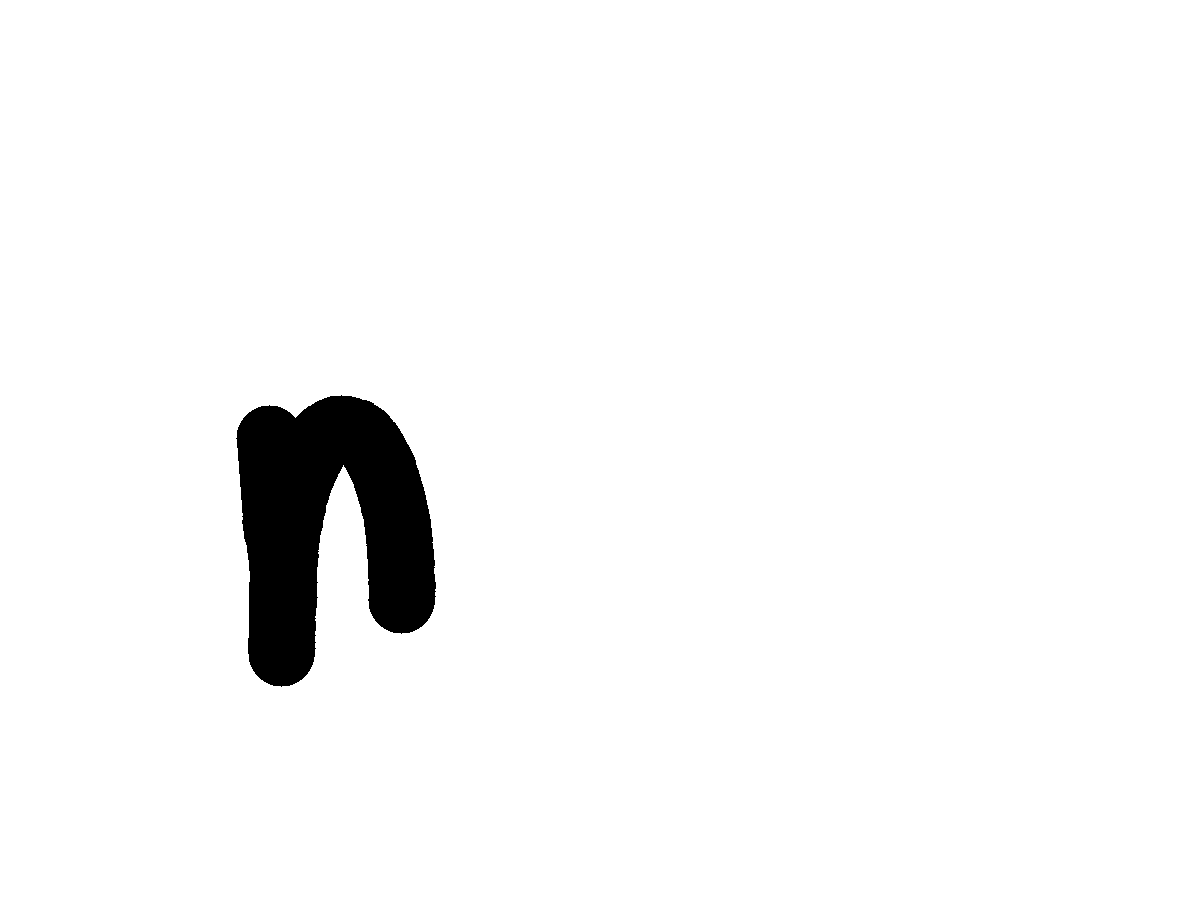}
	\includegraphics[width=0.1\linewidth]{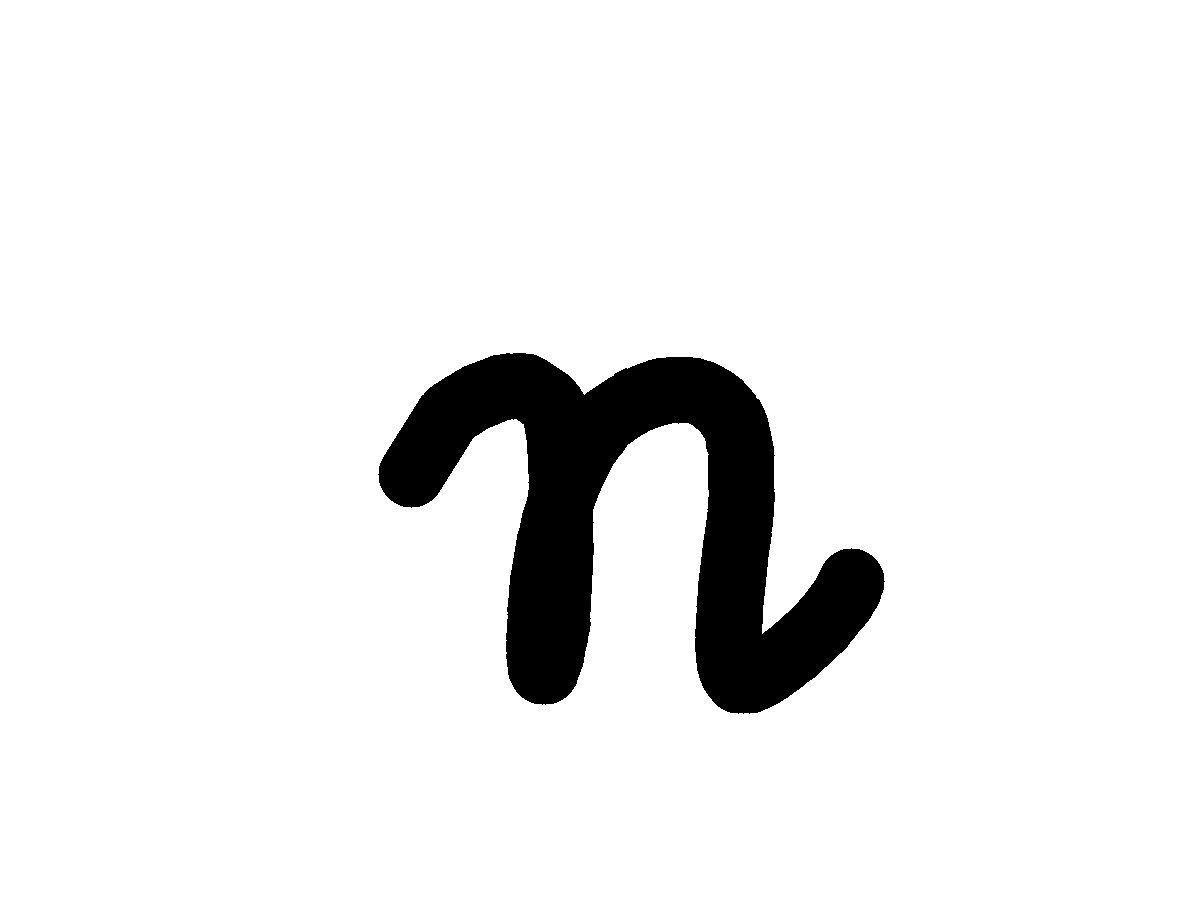}}
\caption{Example data from the character classification dataset. Note that the writing style and typeface varies within each character class. There is also variance in size and position of the characters.}
\label{Figure:character_different}
\end{figure}

\begin{table*}
\centering
\begin{tabular}{c c c c c c c}
\hline
 & Mean & Median & Max & Min & Std & Mean Run Time (s) \\
\hline

kmeans & 55.6\% & 55.0\% & 80.0\% & 36.7\% & 9.8\% & 0.01 \\ 
DTW & 58.4\% & 58.3\% & 85.0\% & 38.3\% & 10.2\% & 0.91 \\ 
LRR & 55.1\% & 55.0\% & 75.0\% & 38.3\% & 9.5\% & 0.17 \\ 
Bayesian & 16.0\% & 0.0\% & 98.3\% & 0.0\% & 34.8\% & 460.87 \\ 
CurveLRR & \hl{84.4\%} & \hl{86.7\%} & \hl{100.0\%} & \hl{45.0\%} & 12.5\% & 291.33 \\ 
\end{tabular}

\caption{Character Trajectory Results}
\label{table_char74k_results}
\end{table*}

In this experiment we used the Chars74K \citep{de2009character} dataset, which consists of pen tip positions (different from trajectories) of handwritten English characters. The dataset consists of 62 character classes with 55 samples per class. Different from the previous English character dataset the writing style or the typeface varies within each character class. For example some of the people choose to close the top of the character ``4'' while others leave it open or they choose to write their characters with differing amounts of serifs as can be seen with the character ``1''. See Figure \ref{Figure:character_different} for visual examples. This creates a significant problem for us as the shape of the data within a class will vary significantly and it is actually best to treat these different font faces as separate classes for greater accuracy. However we leave the dataset's ground truth as is and do not divide the ground truth into separate font faces.

Similar to the previous experiment for each run of this test twenty characters were randomly selected from three random character classes and $50$ runs were performed. However different from the previous experiment we do not apply any further post processing to reduce alignment since this dataset is relatively unprocessed. Results can be found in Table \ref{table_char74k_results}. In spite of the aforementioned challenges with this dataset cLRR once by a significant margin. Notably bayesian clustering suffered many clustering failures leading to very poor accuracy. This was due to implementation limitations in the original code provided from \citep{zhang2015bayesian}.

\subsection{Sign Language Word Clustering}

In this final experiment we use the Australian Sign Language (Auslan) Signs (High Quality) dataset \citep{kadous2002temporal}. This dataset consists of a single native Auslan signer performing 91 different signs (the classes) with 27 samples per sign. The signer wore motion capture gloves that captured the position (x, y, z), roll, pitch and yaw of each hand along with finger bend measurements. This data was captured at $100$Hz and no post processing was applied. The signs were collected three at a time over a period of nine weeks so there is noticeable variation within each class. 

As with previous experiments for each run twenty data points (signed words) were randomly selected from three random word classes and $50$ runs were performed. Only the position, roll, pitch and yaw channels were used since the finger bend measurements were far too noisy and unreliable to be of use. We also performed a parallel test to determine the effectiveness of smoothing. A multi-channel total variation based smoothing method was used \citep{tierney2015selective}. Results can be found in Table \ref{table_asl_results} and \ref{table_asl_results_smooth}. Overall cLRR performed slightly better than Bayesian clustering and significantly better than the baseline methods. We noticed that smoothing the data actually decreased clustering performance of cLRR and LRR but had the opposite effect for all other methods.

\begin{table*}
\centering
\begin{tabular}{c c c c c c c}
\hline
 & Mean & Median & Max & Min & Std & Mean Run Time (s) \\
\hline

kmeans & 64.4\% & 61.7\% & 95.1\% & 40.7\% & 14.8\% & 0.02 \\ 
DTW & 63.8\% & 63.0\% & 92.6\% & 51.9\% & 7.8\% & 0.46 \\ 
LRR & 67.3\% & 66.0\% & 96.3\% & 42.0\% & 14.3\% & 0.27 \\ 
Bayesian & 96.7\% & \hl{100.0\%} & \hl{100.0\%} & 64.2\% & 9.4\% & 186.10 \\ 
CurveLRR & \hl{98.3\%} & \hl{100.0\%} & \hl{100.0\%} & \hl{67.9\%} & \hl{5.4\%} & 98.88 \\ 

\end{tabular}

\caption{Sign Language Results}
\label{table_asl_results}
\end{table*}

\begin{table*}
\centering
\begin{tabular}{c c c c c c c}
\hline
 & Mean & Median & Max & Min & Std & Mean Run Time (s) \\
\hline

kmeans & 58.0\% & 56.8\% & 100.0\% & 0.0\% & 19.6\% & 0.02 \\ 
DTW & 63.3\% & 63.0\% & 92.6\% & 46.9\% & 7.8\% & 0.46 \\ 
LRR & 63.0\% & 62.3\% & 90.1\% & 39.5\% & 12.5\% & 0.27 \\ 
Bayesian & 95.0\% & \hl{100.0\%} & \hl{100.0\%} & 63.0\% & 11.1\% & 186.64 \\ 
CurveLRR & \hl{97.7\%} & \hl{100.0\%} & \hl{100.0\%} & \hl{69.1\%} & \hl{5.6\%} & 98.83 \\ 

\end{tabular}

\caption{Smoothed Sign Language Results}
\label{table_asl_results_smooth}
\end{table*}

\section{Conclusions}
\label{Sec:conclusions}

In this paper, we proposed a tractable algorithm to reliably and accurately cluster functional data and curves. This is a highly challenging problem as functional data from the same class can exhibit stretching, shrinking or be subject to non-uniform warping or scaling. We achieved this by adapting the conventional Euclidean LRR model to a new model over the manifold of open curves. In contrast to other methods our method analyses similarity between data points in their original manifold space, rather than their Euclidean distance. Furthermore in contrast to other methods we gain clustering robustness through low rank regularisation of the global similarity matrix $\mathbf W$, which magnifies the block diagonal structure. Since our method borrows many components of the original LRR model were preserved the computational performance is acceptable and the convergence is guaranteed.

Nevertheless this paper still leaves many areas open for further research. Firstly we only address the data on the manifold of open curves, however much of the data in recognition and computer vision tasks will lie on the manifold of closed curves. Moreover our focus was on relatively clean data with all samples present. In reality some samples can be highly corrupted or missing entirely. We leave solving these problems for a future publication.

\section*{Acknowledgment}

The research project is supported by the Australian Research Council (ARC) through the grant DP140102270. The third author is partly supported by NSFC funded project (No. 41371362) and some of the work was carried out when he was with CSIRO. The fourth author is supported by SAMSI under grant DMS-1127914.

\vskip 0.2in
\bibliography{references}

\end{document}